\documentclass[lettersize,journal]{IEEEtran}

\usepackage{amsmath,amssymb,amsthm,amsfonts}
\usepackage{graphicx}
\usepackage{booktabs}
\usepackage{multirow}
\usepackage{xcolor}
\usepackage[hypertexnames=false]{hyperref}
\usepackage[capitalize]{cleveref}
\usepackage[caption=false,font=normalsize,labelfont=sf,textfont=sf]{subfig}
\usepackage{bm}
\usepackage{enumitem}

\newcommand{\vz}{\mathbf{z}}

\newcommand{\veps}{\bm{\epsilon}}
\newcommand{\sigM}{\sigma_{\!M}}
\newcommand{\seff}{\sigma_{\mathrm{eff}}}
\newcommand{\vrho}{\rho}
\newcommand{\Wtwo}{\mathcal{W}_2}
\providecommand{\xhat}{\hat{x}}
\newcommand{\lamM}{\lambda_{\!M}}
\newcommand{\gM}{g_{\!M}}
\newcommand{\abar}{\bar\alpha}
\newcommand{\Real}{\mathbb{R}}
\newcommand{\Expect}{\mathbb{E}}
\newcommand{\Var}{\mathrm{Var}}
\newcommand{\dd}{\mathrm{d}}
\newcommand{\best}[1]{\textbf{#1}}

\newtheorem{definition}{Definition}
\newtheorem{theorem}{Theorem}
\newtheorem{proposition}{Proposition}

\theoremstyle{remark}
\newtheorem{remark}{Remark}

\title{Improving Diffusion Generative Models via Truncated Karhunen--Lo\`eve Expansion}

\author{Yumeng~Ren, Yaofang~Liu, Aitor~Artola, Laurent~Mertz, Jean-Michel~Morel, and Raymond~H.~Chan%
\thanks{Y.~Ren, Y.~Liu, A.~Artola, and L.~Mertz are with City University of Hong Kong. J.-M.~Morel and R.~H.~Chan are with Lingnan University of Hong Kong.}%
}

\begin{document}
\maketitle

\begin{abstract}
Pretrained diffusion models exhibit a well-known training-sampling mismatch, often attributed to exposure bias and related distribution-shift effects. We provide a quantitative interpretation of this phenomenon through the notion of an effective noise level: empirically, a pretrained denoiser behaves as if trained at a noise level slightly below the nominal schedule. Motivated by this observation, we introduce a training-free sampling strategy based on truncating the Karhunen--Lo\`eve (KL) expansion of the Brownian motion driving the forward stochastic differential equation. 
Truncation yields a finite-dimensional forward process with a reduced noise level that can be adjusted independently of the time discretization. We prove uniform convergence of the truncated process to the original diffusion. To explain the resulting behaviour, we analyse a toy model in which the denoiser is exact but operates at a reduced effective noise level. The analysis predicts a non-monotone response to the sampling noise with a unique interior optimum, located by a one-dimensional sweep over the truncation order. 
We implement the approach through corresponding truncated reverse-time and probability-flow equations, without modifying the network architecture. Across CIFAR-10, CelebA, ImageNet, and latent-space Stable Diffusion, the truncation order consistently reveals a sweet spot, improving pretrained models in nearly all tested configurations. Training from scratch at a matched truncation order makes that order the network's own sweet spot, accelerates convergence by about $2.8\times$, and lowers the generation error on CIFAR-10 (Fr\'echet Inception Distance 7.14 to 5.47 at matched epochs; best checkpoint 6.74 to 5.23). A L\'evy--Ciesielski comparison confirms that finite expansion is broadly beneficial.
\end{abstract}


\section{Introduction}
\label{sec:intro}

Denoising diffusion models~\cite{hoDenoisingDiffusionProbabilistic2020,songScoreBasedGenerativeModeling2021} generate by reversing a fixed forward noising process. Both the theory and standard practice assume the sampler uses the \emph{same} noise schedule $\sigma(t)$ the model was trained for: the reverse process is built from matched forward/reverse marginals, and the learned score is tied to $\sigma(t)$~\cite{songScoreBasedGenerativeModeling2021,linCommonDiffusionNoise2024}. Yet a growing body of work shows this assumption is violated in practice---a finite-capacity denoiser trained with SGD does \emph{not} behave as the exact denoiser at the prescribed level, but at a noise level it has effectively internalized. This mismatch is counterintuitive, given diffusion's reliance on matched marginals, but it is by now well documented under several names: \emph{exposure bias}, corrected training-free by Epsilon Scaling~\cite{ningExposureBiasElucidating2024}; train/sampling schedule defects~\cite{linCommonDiffusionNoise2024}; structured score-estimation error~\cite{caoExploringOptimalChoice2023,hangEfficientDiffusionTraining2023}; and the optimal reverse variance of Analytic-DPM~\cite{baoAnalyticDPMAnalyticEstimate2022}, which differs from the prescribed one. These works document the mismatch and propose empirical or per-step corrections, but leave open a \emph{quantitative} question: at what noise level does a pretrained network actually operate, and what schedule should a sampler use as a result?

\textbf{KL truncation as a noise-level dial.}
To probe a phenomenon about the noise \emph{level} rather than the time grid, we need a knob that lowers the per-time noise magnitude while leaving the time discretization untouched. Our starting point is the variance-preserving (VP) forward SDE~\cite{songScoreBasedGenerativeModeling2021}
\begin{equation}\label{eq:vpsde}
  \dd X_t = f(t)\,X_t\,\dd t + g(t)\,\dd W_t,\qquad X_0 \sim p_{\mathrm{data}},
\end{equation}
with drift $f$ and diffusion $g$. Expanding its Brownian driver $W_t$ in the Karhunen--Lo\`eve (KL) basis and keeping the first $M$ modes yields a finite-dimensional \textbf{KL forward process} $X_t^{(M)}$ with per-time noise variance $\sigM(t)^2 = \alpha(t)^2\sum_{m=1}^{M} c_m(t)^2$, where $\alpha(t)=\exp\!\bigl(\int_0^t f(s) ds\bigr)$ is the signal-scaling factor and the $c_m(t)$ are the driver's modal coefficients (\Cref{sec:parseval}). This level lies strictly below the prescribed $\sigma(t)$ and rises to it as $M\to\infty$ at a uniform, first-order rate, $\sup_t\Expect|X_t-X_t^{(M)}|^2\le C/M$ (\Cref{thm:convergence}, with $M$ the truncation order and $C$ a schedule-dependent constant). Because a DDIM-type sampler accesses the forward process only through its per-time marginal, truncation acts as a \emph{reduced noise schedule} $\sigM(t)<\sigma(t)$ whose contraction $\sigM(t)/\sigma(t)$ varies with $t$, rather than rescaling uniformly. The order $M$ is thus a single noise-level dial, decoupled from the number of sampler steps, and the KL basis is the $L^2$-optimal rank-$M$ choice~\cite{loeve1978}.

\begin{figure*}[t!]
\centering
\includegraphics[width=0.88\textwidth]{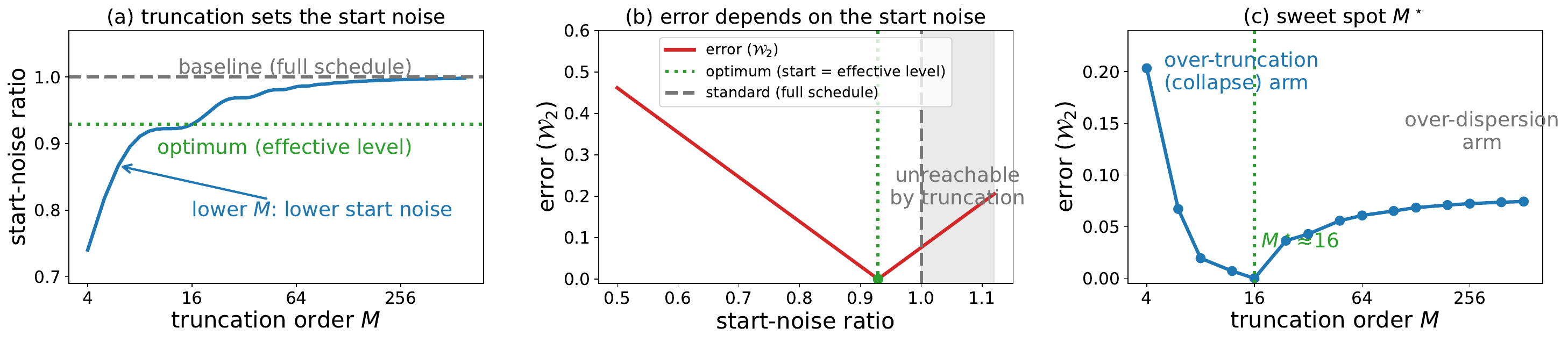}
\caption{The KL-truncation sweet spot: an illustrative computation on the scalar Gaussian model of \Cref{prop:sweetspot} (not measured data); $M$ is the truncation dial and $M^*$ the effective mode count.
The \emph{start-noise ratio} $\sigma^\dagger(t_S)/\sigma(t_S)$ is the level the sampler starts
from, relative to the prescribed one (\Cref{sec:mstar-prediction}, see \eqref{eq:eff-gamma}).
\textbf{(a)} Lowering $M$ lowers this ratio: it rises with $M$ toward the baseline $1$, and one
interior order matches the network's (hidden) effective level. \textbf{(b)} The resulting
generation error---a $2$-Wasserstein surrogate for FID---vanishes when the start noise matches
the effective level and grows on either side; the standard full schedule and the region above
it that truncation cannot reach are marked. \textbf{(c)} Sweeping $M$ therefore traces an
asymmetric U with a single interior minimum at $M^*$. (The effective level is hidden, fixed
here only to place $M^*$; settings: VP schedule, $S\!=\!20$ steps, first step at $t=0.95$ to
avoid the $\alpha\!\approx\!0$ degeneracy at $t=1$.)}
\label{fig:sweetspot}
\end{figure*}

\textbf{Effective noise and the sweet spot.}
To turn the mismatch into a prediction, we model a pretrained denoiser as the exact minimum mean-square-error (MMSE) denoiser of the data at a reduced \emph{effective} noise level $\seff(t)\le\sigma(t)$. This deficiency is expected \emph{a priori}: fitting on finitely many samples, together with the regularization implicit in early-stopped training, shrinks the variance the network reproduces, so the noise it effectively injects is sub-nominal. Sweeping the dial $M$ moves $\sigM$ across $\seff$, and a closed-form $2$-Wasserstein analysis predicts---and our experiments confirm---a \emph{non-monotone} response: as the schedule is lowered, sample quality first \emph{improves}, reaches an interior optimum near the effective level, then degrades back toward the baseline (\Cref{prop:sweetspot}, \Cref{fig:sweetspot}). The answer to the question above is therefore concrete and actionable: a sampler should run \emph{below} the prescribed schedule, at a sweet spot set by the network's own effective level---a training-free gain on fixed checkpoints. We call the order at this optimum the \textbf{effective mode count} $M^*$, and we find the same $M^*$ whether it is located by a sampling-side sweep on pretrained checkpoints or by a training-side sweep.

\textbf{Why KL?}
The sweet spot is a property of the mismatch itself, so any monotonically reduced schedule family exhibits it. The KL advantage is its $t$-dependent contraction: $\sigM(t)/\sigma(t)$ varies with $t$ where a flat rescale cannot. On CIFAR-10 a uniform-rescaling control reproduces the sweet spot but recovers only part of the KL gain (\Cref{sec:rho}); we read this as evidence---not proof---that the shaped contraction matches the network's effective deviation better than a constant one. Finally, the same construction supplies a \textbf{KL reverse-time equation} and its probability-flow counterpart, yielding KL versions of standard deterministic and learned-predictor samplers (KL-DDIM, KL-DPM-Solver) at no extra network cost, with one effective order $M^*$ governing both training and sampling.

\textbf{Contributions.}
\emph{(i)~Theory.} From an exact spectral (Parseval) representation of the VP forward SDE we define the $M$-truncated KL forward process, prove its uniform first-order convergence to the original process, and---modelling a pretrained denoiser at a reduced effective noise level---give a closed-form $2$-Wasserstein analysis (\Cref{prop:sweetspot}) that establishes the non-monotone sweet spot, with finite-sample shrinkage as its \emph{a priori} cause and the optimal order $M^*$ read off empirically.
\emph{(ii)~$M^*$ via sampling (training-free).} The KL schedule family probes $M^*$ on \emph{pretrained} checkpoints, yielding training-free gains across datasets (CIFAR-10, CelebA, ImageNet~$64^2$, Stable Diffusion v1.4) and samplers, from formula-based solvers (DDIM, DPM-Solver(++)) to the learned-predictor solver AMED (\Cref{fig:msweep-universal}), and extends to inverse-problem solving with DAPS.
\emph{(iii)~$M^*$ via training.} A controlled study shows that the sampling-side and training-side sweeps recover the same $M^*$; training at this matched order gives the best pipeline ($2.8\times$ faster convergence and lower FID) and a simple descending-$M^*$ procedure (\Cref{sec:train-scratch}).
\emph{(iv)~Generality.} A L\'evy--Ciesielski comparison shows that finite-dimensional expansion is broadly beneficial, with the $L^2$-optimal KL basis clearly ahead at first order and the two bases comparable at second order.


\section{Related Work}
\label{sec:related}

\paragraph{Denoising Diffusion Models}
DDPM~\cite{hoDenoisingDiffusionProbabilistic2020} showed iterative Gaussian denoising can rival GANs; Song~et~al.~\cite{songScoreBasedGenerativeModeling2021} unified this with score-based models via SDEs, and latent diffusion~\cite{rombach2022high} and classifier-free guidance~\cite{ho2022classifier} scaled it to high-resolution conditional synthesis. Modified forward processes have also been explored---blurring/heat dissipation~\cite{hoogeboom2023blurring}, arbitrary degradations~\cite{bansal2023cold}, heavy-tailed L\'evy drivers~\cite{yoon2023levy}, and fractional Brownian drivers~\cite{nobis2024fractional}---but these replace the driving process or the degradation itself. To our knowledge, we are the first to modify the \emph{spectral content} of the standard Brownian driver by finite-dimensional KL truncation, leaving the VP process family otherwise unchanged.

\paragraph{Accelerated Sampling}
DDIM~\cite{songDenoisingDiffusionImplicit2021} recast the reverse process as a non-Markovian chain; DPM-Solver(++)~\cite{lu2022dpm,lu2022dpmpp} apply exponential integrators to the probability-flow ODE; EDM~\cite{karrasElucidatingDesignSpace2022} tunes schedule and preconditioning; Analytic-DPM~\cite{baoAnalyticDPMAnalyticEstimate2022} gives closed-form optimal per-step variance; and distillation~\cite{salimans2022progressive}, consistency models~\cite{song2023consistency}, rectified flows~\cite{liuFlowStraightFast2022b}, and AMED~\cite{zhou2024amed} cut evaluations via extra training. All operate on the standard forward process; our KL samplers add a complementary, training-free dimension---they alter the noise magnitude at fixed discretization, while these methods accelerate the discretization itself.

\paragraph{Score Estimation Error and Schedule Mismatch}
That a trained denoiser operates at a noise level different from the prescribed one is documented as structured score error~\cite{caoExploringOptimalChoice2023,hangEfficientDiffusionTraining2023}, exposure bias / Epsilon Scaling~\cite{ningExposureBiasElucidating2024}, train/sampling schedule defects~\cite{linCommonDiffusionNoise2024}, and the optimal reverse variance of Analytic-DPM~\cite{baoAnalyticDPMAnalyticEstimate2022}. These are empirical or per-step corrections of the operating point; we instead give a closed-form account of the discrepancy and its interior optimum via a one-parameter KL schedule family $\{\sigM(t)\}$ (\Cref{prop:sweetspot}). Unlike schedule-optimization methods that tune the time-discretization for a correctly specified score~\cite{karrasElucidatingDesignSpace2022,sabourAlignYourSteps2024,elementaryScheduling2026}, we fix the discretization and vary the noise \emph{magnitude} $\sigM(t)<\sigma(t)$ at fixed $t$---a distinct degree of freedom.

\paragraph{Karhunen--Lo\`eve Expansion in Computational Science}
The KL expansion is a classical, $L^2$-optimal spectral decomposition of stochastic processes~\cite{loeve1978,pavliotisStochasticProcessesApplications2014}, widely used in stochastic finite elements~\cite{ghanem2003stochastic}, uncertainty quantification, random-field generation~\cite{schwab2006karhunen}, and inverse problems~\cite{kawarSNIPSSolvingNoisy2021}. To our knowledge this is the first use of the KL expansion of the Brownian motion \emph{driving the forward SDE} to build a finite-dimensional diffusion framework.

\section{The KL-Diffusion Framework}
\label{sec:method}

We develop the KL-Diffusion framework in full.
Section~\ref{sec:parseval} derives the exact Parseval representation.
Section~\ref{sec:kl-diffusion} introduces the truncated KL forward
process.
Section~\ref{sec:kl-matrix} recasts mode truncation as a reduced noise
schedule---the noise-level dial.
Section~\ref{sec:mstar-prediction} gives the exact Gaussian analysis whose
central prediction is an \emph{effective mode count}~$M^*$.
Section~\ref{sec:kl-reverse} derives the KL reverse-time equation,
Section~\ref{sec:kl-loss} the KL training loss (with training mode count~$M'$),
and Section~\ref{sec:kl-samplers} the KL sampler family (KL-DDIM, KL-Euler,
KL-DPM-Solver).
This prediction is then measured empirically in the experiments:
\Cref{sec:train-scratch} confirms~$M^*$ from both the sampling and training
axes on CIFAR-10, and \Cref{sec:pretrained-samplers} probes it across datasets,
samplers, and model scales.

\subsection{KL Expansion of the SDE Solution}
\label{sec:parseval}

We work on $[0,T]$ and adopt the variance-preserving (VP) forward
SDE~\eqref{eq:vpsde}, with drift $f(t)=-\tfrac12\beta(t)$ and diffusion
$g(t)=\sqrt{\beta(t)}$, where $\beta(t) = \beta_{\min} + (\beta_{\max} -
\beta_{\min})t/T$ and $\beta_{\max}>\beta_{\min}>0$.
Its solution is
\begin{equation}\label{eq:SDEsoln}
  X_t = \alpha(t)\,X_0 + \alpha(t)\!\int_0^t h(s)\,\dd W_s,
\end{equation}
with signal-scaling factor $\alpha(t) =
\exp\!\bigl(-\tfrac{1}{2}\!\int_0^t\!\beta(s)\,\dd s\bigr)$,
reduced diffusion coefficient $h(s) = \sqrt{\beta(s)}/\alpha(s)$, and
marginal variance $\sigma(t)^2 = 1 - \alpha(t)^2$.

The Brownian motion on $[0,T]$ admits the KL
expansion~\cite{pavliotisStochasticProcessesApplications2014} with
eigenfunctions $\varphi_m(t) = \sqrt{2/T}\sin\!\bigl((m\!-\!\tfrac{1}{2})
\pi t/T\bigr)$ and eigenvalues
$\lambda_m = T^2/\bigl((m\!-\!\tfrac{1}{2})^2\pi^2\bigr)$.
The \textbf{derivative eigenfunctions}
\begin{equation}\label{eq:psi}
  \psi_m(t) := \sqrt{\lambda_m}\,\varphi_m'(t)
  = \sqrt{\tfrac{2}{T}}\cos\!\Bigl(\bigl(m\!-\!\tfrac{1}{2}\bigr)\tfrac{\pi t}{T}\Bigr)
\end{equation}
form a complete orthonormal basis of $L^2([0,T])$.
By the Wiener isometry~\cite{oksendal2003}, defining
$Z_m := \int_0^T \psi_m(s)\,\dd W_s$, we have
$Z_m \overset{\mathrm{iid}}{\sim} \mathcal{N}(0,1)$, and Parseval's
theorem yields the \textbf{exact representation} of the SDE solution \eqref{eq:SDEsoln}:
\begin{equation}\label{eq:exact}
  X_t = \alpha(t)\,X_0 + \alpha(t)\sum_{m=1}^{\infty} c_m(t)\,Z_m,
\end{equation}
where $c_m(t) := \int_0^t h(s)\,\psi_m(s)\,\dd s$.
This is exact (not an approximation) and forms the foundation of our
framework: the randomness of the SDE solution can be reorganized into
independent KL modes before any truncation is introduced.

\subsection{The Truncated KL Forward Process}
\label{sec:kl-diffusion}

Retaining only the first $M$ terms in~\eqref{eq:exact} defines the
\textbf{KL forward process}:
\begin{equation}\label{eq:XM}
\begin{aligned}
  X_t^{(M)} &:= \alpha(t)\Bigl(X_0 + \sum_{m=1}^{M} Z_m\,c_m(t)\Bigr),\\
  \vz&=(Z_1,\dots,Z_M)^{\top} \sim \mathcal{N}(\mathbf{0}, I_M).
\end{aligned}
\end{equation}
This is the $M$-mode approximation of \eqref{eq:SDEsoln} or
\eqref{eq:exact}, with augmented Gaussian vector $\vz$ and deterministic
temporal coefficients $c_m(t)$.
The marginal variance satisfies
\begin{equation}\label{eq:sigM}
  \sigM(t)^2 := \alpha(t)^2 \sum_{m=1}^{M} c_m(t)^2
  \xrightarrow{M\to\infty} \sigma(t)^2,
\end{equation}
by Parseval's theorem.

\begin{theorem}[Convergence]\label{thm:convergence}
Assume $h \in C^1([0,T])$.  Then
$\sup_{t}\,\Expect[|X_t - X_t^{(M)}|^2] \leq C/M$,
where $C = (\max_t \alpha(t))^2 \cdot A^2 / \pi^2$ with
$A = \sqrt{2}(\|h\|_\infty + \|h'\|_\infty)$.
\end{theorem}

The verification (Appendix~\ref{app:convergence}, stated with the normalization $T{=}1$; general $T$ follows by time rescaling) exploits the fact that
$X_t - X_t^{(M)} = \alpha(t)\sum_{m>M} c_m(t)\,Z_m$ is a Parseval tail,
with $|c_m(t)| \leq A/((m\!-\!\tfrac{1}{2})\pi)$ by integration by parts.

\begin{remark}[Random-ODE form]
Differentiating the truncated solution \eqref{eq:XM} gives the equivalent
forward equation
\begin{equation}\label{eq:kl-forward-sde}
  \dd X_t^{(M)} = f(t)\,X_t^{(M)}\,\dd t + g(t)\,\dd W_t^{(M)},
\end{equation} where
$W_t^{(M)} := \sum_{m=1}^{M} Z_m\,\sqrt{\lambda_m}\,\varphi_m(t)$ is the truncated KL Brownian motion.
This is a random ODE with augmented Gaussian initial vector $\vz$ and
deterministic temporal coefficients $\varphi_m(t)$. It takes \eqref{eq:XM} as the unique solution.
\end{remark}

The convergence result shows that the truncated KL forward process
recovers the VP SDE as $M$ grows, but it does not yet explain what a
finite value of $M$ means for a discretized diffusion trajectory.
We next make this relation explicit through a linear system connecting
the latent KL modes to time observations.

\subsection{From Mode Truncation to a Reduced Noise Schedule}
\label{sec:kl-matrix}

The truncated marginal variance \eqref{eq:sigM},
$\sigM(t)^2=\alpha(t)^2\sum_{m\le M}c_m(t)^2$, is non-decreasing in $M$ and increases to
$\sigma(t)^2$ as $M\to\infty$ (\Cref{thm:convergence}). Because the KL driver is white
noise, truncation lowers the variance of each per-time marginal \emph{without} introducing
non-Gaussian or cross-time structure, and a DDIM-type sampler accesses the forward process
only through this per-time marginal level. The sampler therefore cannot distinguish the
order-$M$ truncated process from the full process run with the reduced schedule $\sigM(t)$:
\begin{quote}
\emph{mode truncation acts, at the sampler, as a reduced noise schedule
$\sigM(t)<\sigma(t)$ whose contraction $\sigM(t)/\sigma(t)$ varies with $t$.}
\end{quote}
The truncation order~$M$ is thus a dial on the sampling noise level, decoupled from the
time discretization; the $t$-dependence of $\sigM/\sigma$---as opposed to a uniform
rescaling---is what later distinguishes the KL family from a constant-scaled schedule
(\Cref{sec:rho}).

\subsection{Exact Gaussian Analysis and the Sampling Sweet Spot}
\label{sec:mstar-prediction}

Under an isotropic Gaussian prior the score is linear, the Bayes-optimal denoiser is a
Wiener filter, and the deterministic ($\eta\!=\!0$) DDIM chain collapses to a product of
scalars. This yields an exact account of the sampled variance and of \emph{why} a
truncation sweet spot exists. Throughout we order the schedule in reverse time
$1=t_S>\cdots>t_0=0$ ($t_S$ the noise end, $t_0$ the data end), with $S$ the number of
sampler steps.

\paragraph{Closed-form estimators.}
Take the scalar Gaussian core $X_0\sim\mathcal{N}(0,\gamma^2)$ (the per-coordinate prior;
all results extend coordinate-wise). With $X_t=\alpha(t)X_0+\sigma(t)\veps$
(\Cref{sec:parseval}), the marginal of $X_t$ is $\mathcal{N}(0,\vrho^2(t))$ with total
marginal variance
\begin{equation}\label{eq:varrho}
  \vrho^2(t):=\alpha(t)^2\gamma^2+\sigma(t)^2 .
\end{equation}
The optimal (MMSE) noise predictor and clean-data estimator are the Wiener filters
\begin{equation}\label{eq:wiener}
  \veps^*(x,t)=\frac{\sigma(t)}{\vrho^2(t)}\,x,\quad
  \xhat_0^*(x,t)=\Expect[X_0\mid X_t\!=\!x]=\frac{\alpha(t)\gamma^2}{\vrho^2(t)}\,x,
\end{equation}
both linear in $x$, with score $\nabla_x\log p_t(x)=-x/\vrho^2(t)$. These are the standard
Gaussian/MMSE score and denoiser, which we cite rather than
re-derive~\cite{wangVastola2024,hurault2024,vincent2011,efron2011}.

\paragraph{Scalar transfer coefficient and polar form.}
Let $\sigma^\dagger(t)$ denote the noise schedule used by the sampler, in the deterministic DDIM
step and to set the start point, with the forward estimators \eqref{eq:wiener} unchanged.
Substituting \eqref{eq:wiener} into the $\eta\!=\!0$ update
$X_s^\dagger=\alpha(s)\xhat_0^*(X_t^\dagger,t)+\sigma^\dagger(s)\veps^*(X_t^\dagger,t)$ makes each
reverse step a scalar multiplication $X_s^\dagger=A^\dagger(t,s)\,X_t^\dagger$ ($s<t$), with
\begin{equation}\label{eq:Ats}
  A^\dagger(t,s)=\frac{\alpha(s)\alpha(t)\gamma^2+\sigma^\dagger(s)\,\sigma(t)}
                     {\alpha(t)^2\gamma^2+\sigma(t)^2}.
\end{equation}
The numerator is an
inner product; with the \emph{data-scale diffusion angles}
\begin{equation}\label{eq:phi-angle}
  \phi(t):=\arctan\frac{\sigma(t)}{\alpha(t)\gamma},\qquad
  \phi^\dagger(t):=\arctan\frac{\sigma^\dagger(t)}{\alpha(t)\gamma}\;\in\Big[0,\tfrac{\pi}{2}\Big),
\end{equation}
and $\vrho^\dagger(s):=\sqrt{\alpha(s)^2\gamma^2+\sigma^\dagger(s)^2}$---so that
$\alpha(s)\gamma=\vrho^\dagger(s)\cos\phi^\dagger(s)$, $\sigma^\dagger(s)=\vrho^\dagger(s)\sin\phi^\dagger(s)$
and $\alpha(t)\gamma=\vrho(t)\cos\phi(t)$, $\sigma(t)=\vrho(t)\sin\phi(t)$---the cosine addition
formula gives $A^\dagger(t,s)=\big(\vrho^\dagger(s)/\vrho(t)\big)\cos\!\big(\phi(t)-\phi^\dagger(s)\big)$.
Folding the data scale $\gamma$ into the angle \emph{generalizes} the angular DDIM
parameterization of Salimans and Ho~\cite{salimans2022progressive}.

\paragraph{Output scale of the $S$-step chain.}
Over the $S$ steps the scalar transfers multiply, so the chain maps its start
state to
$X_{t_0}^\dagger=\bigl(\prod_{k=1}^{S}A^\dagger(t_k,t_{k-1})\bigr)X_{t_S}^\dagger$,
and the object to estimate is the transfer product. We approximate it in two
steps. \emph{(i)}~Inside each transfer coefficient we set
$\sigma^\dagger\approx\sigma$, hence $\vrho^\dagger\approx\vrho$ and
$\phi^\dagger\approx\phi$; by the polar form,
\begin{align}\label{eq:Sprod}
  \prod_{k=1}^{S}A^\dagger(t_k,t_{k-1})
  &\approx\prod_{k=1}^{S}\frac{\vrho(t_{k-1})}{\vrho(t_k)}
   \cos\bigl(\phi(t_k)-\phi(t_{k-1})\bigr)\notag\\
  &=\frac{\gamma}{\vrho(t_S)}\prod_{k=1}^{S}\cos\bigl(\phi(t_k)-\phi(t_{k-1})\bigr),
\end{align}
where the $\vrho$-ratios telescope and
$\vrho(t_0)=\sqrt{\alpha(0)^2\gamma^2+\sigma(0)^2}=\gamma$.
\emph{(ii)}~The remaining cosine product depends on the grid only through the
angle increments, which are nonnegative and sum to the total sweep
$\Phi=\phi(t_S)-\phi(t_0)=\phi(t_S)$. We replace it by its best case over
grids, the \emph{discretization factor}
\begin{equation}\label{eq:kappa}
  \kappa_S:=\sup\Big\{\prod_{k=1}^{S}\cos\delta_k\ :\ \delta_k\ge0,\ \sum_{k=1}^{S}\delta_k=\Phi\Big\}\;\in(0,1),
\end{equation}
attained at equal angles, $\kappa_S=\cos^S(\Phi/S)\approx e^{-\Phi^2/2S}$
(Appendix~\ref{app:uniform-angle}); the product for any actual grid is smaller by an
$O(1/S)$ factor, an uncertainty inherited by all downstream uses of $\kappa_S$.
Thus $\prod_{k=1}^{S}A^\dagger(t_k,t_{k-1})\approx\gamma\kappa_S/\vrho(t_S)$.
\emph{(iii)}~Finally, the chain is linear and is initialised by the sampling
schedule at
$X_{t_S}^\dagger\sim\mathcal N\bigl(0,(\sigma^\dagger(t_S))^2 I_d\bigr)$, so the
output is a centred Gaussian whose scale is approximated by
\begin{equation}\label{eq:eff-gamma}
  \sigma^\dagger(t_S)\prod_{k=1}^{S}A^\dagger(t_k,t_{k-1})
  \;\approx\;\frac{\gamma\,\kappa_S}{\vrho(t_S)}\,\sigma^\dagger(t_S)
  \;=:\;\tilde\gamma_S^\dagger,
\end{equation}
i.e.\ the output is approximately $\mathcal{N}(0,(\tilde\gamma_S^\dagger)^2)$;
$\tilde\gamma_S^\dagger$ names the approximant on the right of
\eqref{eq:eff-gamma}, not the exact output scale. After steps (i)--(ii) the
sampling schedule enters the output only through its start value
$\sigma^\dagger(t_S)$.

\paragraph{Effective forward noise.}
A finite-capacity network does not realize the prescribed forward noise exactly; within the
KL family this under-realization is indexed by a finite order, i.e.\ a reduced schedule
$\sigM(t)$ (indeed, \Cref{sec:kl-loss} shows that KL training at order $M'$ is exactly
training at the reduced schedule $\sigma_{M'}$). We therefore posit, as a modeling assumption corroborated by the experiments, an \emph{effective}
forward level $\seff(t)\le\sigma(t)$ that is deficient at the start, $\seff(t_S)<\sigma(t_S)$.
This is the level the estimators \eqref{eq:wiener} effectively use, so the prefactor
$\gamma/\vrho(t_S)$ in \eqref{eq:eff-gamma} is read at it,
$\vrho(t_S)\!\to\!\vrho_{\mathrm{eff}}(t_S)\approx\seff(t_S)$, while the start value
$\sigma^\dagger(t_S)$ set by the sampling schedule is untouched:
\begin{equation}\label{eq:tilde-corr}
  \tilde\gamma_S^\dagger\approx\gamma\,\kappa_S\,\frac{\sigma^\dagger(t_S)}{\seff(t_S)} .
\end{equation}
The two levels now play distinct roles: $\sigma^\dagger(t_S)$ (numerator) is the inference
noise the sampler may choose, whereas $\seff(t_S)$ (denominator) is what the network
implements; the truncated KL schedule makes $\seff(t_S)$ concrete through the order~$M$.

Deficiency is expected \emph{a priori} from finite-sample fitting and the regularization
implicit in early-stopped training (\Cref{sec:intro}); the same prescribed-versus-effective
discrepancy is reported empirically as exposure bias / noise-level correction
(\Cref{sec:related}).

\begin{definition}[Effective noise level]
\label{def:eff}
A denoiser $\xhat_0^\theta(\cdot,t)$ has \emph{effective noise level} $\seff(t)\le\sigma(t)$
if it equals the exact MMSE denoiser of the data at additive level $\seff(t)$,
$\xhat_0^\theta(x,t)=\Expect[X_0\mid \alpha(t)X_0+\seff(t)\,\xi=x]$,
$\xi\sim\mathcal{N}(0,I)$; it is \emph{deficient} when $\seff(t)<\sigma(t)$.
\end{definition}

Measuring fidelity with the
$2$-Wasserstein distance between centred isotropic Gaussians,
$\Wtwo(\mathcal{N}(0,a^2I_d),\mathcal{N}(0,b^2I_d))=\sqrt d\,|a-b|$---a deliberate,
network-free surrogate for FID that captures exactly the second-moment fidelity the Gaussian
model controls---yields the sweet spot.

\begin{proposition}[Sampling sweet spot]
\label{prop:sweetspot}
With $D:=\Wtwo\big(\mathcal{N}(0,(\tilde\gamma_S^\dagger)^2 I_d),\mathcal{N}(0,\gamma^2 I_d)\big)$,
\begin{equation}\label{eq:parabola}
\begin{aligned}
  D^2 &= d\,\gamma^2\Big(\kappa_S\,\frac{\sigma^\dagger(t_S)}{\seff(t_S)}-1\Big)^2,\\
  \text{optimum:}\quad \sigma^\dagger(t_S) &= \frac{\seff(t_S)}{\kappa_S} .
\end{aligned}
\end{equation}
Since $D$ is the modulus of an affine function of $\sigma^\dagger(t_S)$, this zero is its unique
global minimum. The optimal start noise sits a factor $1/\kappa_S>1$ above the effective level: it
\emph{amplifies} ($\sigma^\dagger(t_S)>\sigma(t_S)$) when $\seff(t_S)>\kappa_S\sigma(t_S)$ and
\emph{attenuates} ($\sigma^\dagger(t_S)<\sigma(t_S)$) otherwise---one formula, both regimes.
\end{proposition}

Because $\seff(t_S)<\sigma(t_S)$ is assumed from the start, both regimes arise without any
contradiction-then-fix; the image experiments fall in the attenuating branch (sampling below
the prescribed schedule improves fidelity). The polar/tangent device coincides with that used
to optimize schedules for a correctly specified score~\cite{sabourAlignYourSteps2024}; here
the discretization is fixed. \Cref{prop:sweetspot} thus establishes that the sweet spot
\emph{exists}, is \emph{unique}, and is \emph{locatable by a one-dimensional sweep}; through
the $\kappa_S$ reduction it is deliberately blind to the numerical value of $M^*$ and to which
reduced family is best---both are read from experiments. Note that if $\seff=\sigma$ the
optimum $\sigma(t_S)/\kappa_S$ would lie \emph{above} the prescribed schedule and every
reduced schedule would only degrade quality, contrary to \Cref{fig:msweep-universal}: the
deficiency $\seff<\sigma$ is the minimal modification consistent with the data. Note also that
$\seff$ replaces the \emph{estimator-side} level in \eqref{eq:wiener}, not the sampler-side
$\sigma^\dagger$, so within the model it is a checkpoint property.

\paragraph{Realization by the KL schedule.}
A physically motivated choice of the sampling schedule is the KL schedule, $\sigma^\dagger=\sigM$
(\Cref{sec:kl-diffusion}), realized in practice by KL-DDIM (\Cref{sec:kl-samplers}). Only the
start value enters, so
\begin{equation}\label{eq:D-M}
  D(M)=\gamma\sqrt d\,\Big|\kappa_S\,\frac{\sigM(t_S)}{\seff(t_S)}-1\Big| ,
\end{equation}
and as $M$ increases $\sigM(t_S)\uparrow\sigma(t_S)$, so $\sigM(t_S)$ sweeps upward and crosses
$\seff(t_S)/\kappa_S$ at an interior order $M^*$: $D(M)$ first falls to zero there and then
rises---a single-minimum U-shape. The order at the optimum is the network's \textbf{effective
mode count}~$M^*$, located per checkpoint--sampler configuration; the sampling-side and
training-side probes cohere at the fixed point $M^*\!=\!M'$ (\Cref{sec:train-scratch}). As FID
is itself a Bures $\Wtwo^2$ between Gaussian feature fits, the structural predictions---steep
collapse arm, interior optimum, gentle approach to the baseline---transfer to the FID curve,
though absolute values are not.

\paragraph{Beyond the $\kappa_S$ reduction.}
\Cref{prop:sweetspot} treats $\sigma^\dagger$ as unknown, and $\kappa_S$ is the price: the
U-shape it delivers is governed by the start values $\sigma(t_S)$ and $\sigma^\dagger(t_S)$
alone, not by the profiles $\sigma(t)$ and $\sigma^\dagger(t)$. Once a family
$\{\sigma^\dagger_M\}$ is fixed, one mild assumption on it removes $\kappa_S$; $\seff$ remains
unknown but fixed. Keeping the estimators \eqref{eq:wiener} at the effective level
(\Cref{def:eff}), the $\eta\!=\!0$ update is again a scalar multiplication, now with
\begin{equation}\label{eq:Ats-exact}
  A^\dagger(t,s)=\frac{\alpha(s)\alpha(t)\gamma^{2}+\sigma^\dagger(s)\,\seff(t)}
                      {\alpha(t)^{2}\gamma^{2}+\seff(t)^{2}} ,
\end{equation}
so a chain started at scale $s_0$ has output scale and error
\begin{equation}\label{eq:Pm}
  s(M)=s_0\prod_{k=1}^{S}A^\dagger(t_k,t_{k-1}),\qquad
  D(M)=\sqrt d\,\bigl|s(M)-\gamma\bigr| .
\end{equation}

\begin{proposition}[Sweet spot without the $\kappa_S$ reduction]
\label{prop:sweetspot-exact}
Fix $\seff$ on the sampler grid, with $\seff(t_k)>0$ for some $k\ge1$. If
$\sigma^\dagger_M(t_k)\le\sigma^\dagger_{M'}(t_k)$ at every grid point whenever $M<M'$, with
strict inequality at one of them, then $s(M)$ is strictly increasing in $M$. Hence $D(M)$
falls, then rises: it has a unique minimizer, interior whenever $s(M)$ crosses $\gamma$ on the
swept range.
\end{proposition}

Indeed, each $A^\dagger(t_k,t_{k-1})>0$ and depends on $M$ only through
$\sigma^\dagger_M(t_{k-1})$ in its numerator, and $D$ is the modulus of an affine function of
$s(M)$. Since $s_0$ does not depend on $M$, the start convention is immaterial to the shape: our
samplers initialize from $X_{t_S}\sim\mathcal{N}(0,I_d)$, as standard DDIM does. The KL,
$\nu$- and L\'evy--Ciesielski families all meet the hypotheses of both propositions, so each
exhibits the U-shape; this is the statement the experiments of
\Cref{sec:train-scratch,sec:pretrained-samplers,sec:rho} test. The KL reverse-time equation,
training loss, and samplers are developed next.

\subsection{KL Reverse-Time Equation}
\label{sec:kl-reverse}

Since the truncated KL forward process is a random ODE rather than an
It\^o SDE, we construct its reverse-time counterpart indirectly: we first
introduce an auxiliary It\^o SDE whose marginals match those of the KL forward
process in the $M\to\infty$ limit (exact matching is generally impossible at
finite $M$; Supplementary App.~H), apply Anderson's reversal to
that SDE, and then truncate the reverse noise in the KL basis. The result is a
principled \emph{approximate} reversal, exact as $M\to\infty$; among our
samplers only KL-Euler depends on it, while KL-DDIM and KL-DPM-Solver are
defined directly from the exact per-time marginals of \eqref{eq:XM}
(Supplementary App.~H).
Given terminal state $X_T^{(M)}$ and fresh expansion coefficients
$\{\bar{Z}_m\}_{m=1}^M \overset{\mathrm{iid}}{\sim} \mathcal{N}(0,1)$,
the \textbf{KL reverse-time equation} is:
\begin{align}
  \frac{\dd X_t}{\dd t}
  &= f(t)\,X_t
  - \gM^2(t)\,\nabla_X \log p_t^{(M)}(X_t) \nonumber\\
  &\quad + \gM(t)\sum_{m=1}^{M} \psi_m(t)\,\bar{Z}_m .
  \label{eq:kl-reverse}
\end{align}
where $\nabla_X \log p_t^{(M)}(X_t)$ is the score function for the KL forward equation \eqref{eq:kl-forward-sde}, $\gM(t) = \sqrt{\beta(t)}\,\sigM(t)/\sigma(t)$ is the
\emph{effective diffusion coefficient} of the truncated process, and
$\psi_m(t)$ are the cosine basis functions from~\eqref{eq:psi}.
The fresh coefficients
$\{\bar{Z}_m\}_{m=1}^{M} \overset{\mathrm{iid}}{\sim}
\mathcal{N}(0,1)$ are drawn once and held fixed for the entire
reverse trajectory.
Once drawn, the third term---the \emph{expansion force}---becomes a
known smooth function of~$t$, reducing~\eqref{eq:kl-reverse} to a
deterministic ODE amenable to standard high-order solvers.
The $\{\bar{Z}_m\}$ are independent of the forward $\{Z_m\}$;
Anderson's time-reversal theorem~\cite{anderson1982} and the
Fokker--Planck constraint require only equality in distribution, not
sample-level correspondence.

\begin{remark}[Connection to samplers]
Existing samplers discretize the continuous reverse SDE.
By starting from the KL reverse-time equation~\eqref{eq:kl-reverse}
instead, we obtain a family of KL samplers whose coefficients naturally
involve $\sigM(t)$ rather than $\sigma(t)$.
\end{remark}

\subsection{KL Training Loss}
\label{sec:kl-loss}

The KL forward process~\eqref{eq:XM} directly induces a training loss. Write $M'$ for the
training truncation order (the \textbf{training mode count}; standard training is
$M'\!=\!\infty$) and $M$ for the sampler order. Drawing $X_0\sim p_{\mathrm{data}}$,
$\vz\sim\mathcal{N}(0,I_{M'})$, and $t$, the noised sample is
$X_t^{(M')}=\alpha(t)X_0+\sigma_{M'}\!(t)\hat\veps$ with
$\hat\veps=\alpha(t)\sum_{m=1}^{M'}Z_m c_m(t)/\sigma_{M'}\!(t)$; by additivity of independent
Gaussians $\hat\veps\sim\mathcal{N}(0,1)$ exactly, so for a single-time marginal the KL loss
\begin{equation}\label{eq:kl-loss}
  \mathcal{L}_{\mathrm{KL}}=\Expect_{t,X_0,\vz}\big[\|\veps_\theta(X_t^{(M')},t)-\hat\veps\|^2\big]
\end{equation}
is implemented by drawing $\veps\sim\mathcal{N}(0,I)$ and scaling by $\sigma_{M'}\!(t)$; it
differs from the standard loss only in the reduced noise level and recovers it as
$M'\!\to\!\infty$. We adopt this single-time loss and leave multi-time KL objectives to future
work.

\subsection{KL Samplers}
\label{sec:kl-samplers}

The KL reverse-time equation~\eqref{eq:kl-reverse} provides a principled
starting point for deriving samplers of increasing sophistication.
Throughout the experiments, $S$ denotes the number of sampler steps. We
present the canonical \textbf{KL-DDIM} sampler first---the workhorse of our
experiments---followed by KL-Euler and the higher-order KL-DPM-Solver variants.

\subsubsection{KL-DDIM\texorpdfstring{$(M, \eta)$}{(M, eta)}}
KL-DDIM is the canonical KL sampler used throughout our experiments: it is
the popular deterministic DDIM solver, the order-1 base case to which the
higher-order KL solvers reduce, and its clean $(M,\eta)$ knobs make it the
natural vehicle for both the $S$-sweep and the $M$-sweep.
KL-DDIM is derived from the KL probability flow ODE---the
deterministic counterpart of~\eqref{eq:kl-reverse} obtained by dropping
the expansion force and halving the score coefficient:
\[
  \dot{X}_t = f(t)\,X_t - \tfrac{1}{2}\gM^2(t)\,\nabla_X \log p_t^{(M)}(X_t).
\]
The resulting update is identical to standard
DDIM~\cite{songDenoisingDiffusionImplicit2021} with the replacement
$\sigma(\cdot) \to \sigM(\cdot)$ in both the predicted~$\hat{X}_0$ and
the directional step:
\begin{align}\label{eq:kl-ddim}
  \hat{X}_0 &= \bigl(X_s - \sigM(s)\,\veps_\theta(X_s,s)\bigr)
    / \sqrt{\abar_s}, \notag\\
  X_t &= \sqrt{\abar_t}\,\hat{X}_0
    + \sqrt{\sigM^2(t) - \tilde\sigma_\eta^2}\;\veps_\theta(X_s,s)
    + \tilde\sigma_\eta\,\veps,
\end{align}
where $\tilde\sigma_\eta^2 = \eta^2\,\sigM^2(t)\,
\bigl(1 - \abar_s\,\sigM^2(t)/(\abar_t\,\sigM^2(s))\bigr)$, with $\abar_t = \alpha(t)^2$ as before.
At $\eta = 0$, the sampler is purely deterministic, and
$\sigM(t) < \sigma(t)$ reduces the directional coefficient,
yielding a \emph{less aggressive} correction at each step.
Because KL-DDIM is derived from the KL probability flow ODE---whose
coefficients involve $\sigM(t)$ through the marginal variance of the
KL forward process---the schedule $\sigM(t)$ enters the sampler as a
consequence of the derivation, not as a heuristic substitution.
The resulting implementation differs from standard DDIM only in
replacing $\sigma(t)$ with the precomputed $\sigM(t)$.
Equivalently, and without reference to the reverse-time equation, KL-DDIM is
exactly the non-Markovian DDIM construction
of~\cite{songDenoisingDiffusionImplicit2021} applied to the per-time marginals
$\mathcal N\bigl(\alpha(t)X_0,\sigM^2(t)I\bigr)$ of the KL forward process,
which \eqref{eq:XM} supplies in closed form. All KL samplers initialize from
$X_{t_S}\sim\mathcal N(0,I)$, as standard DDIM does; by
\Cref{prop:sweetspot-exact} the start convention does not affect the shape of
the $M$-response.
For $\eta > 0$, the same KL-DDIM formula injects fresh i.i.d.\
noise $\veps \sim \mathcal{N}(0,I)$ scaled by $\tilde\sigma_\eta$,
exactly as in the stochastic branch of standard DDIM; no separate
sampler is introduced for this case.

\subsubsection{\texorpdfstring{KL-Euler$(M)$}{KL-Euler(M)}}
The most direct approach is to apply the Euler method to the KL
reverse-time equation~\eqref{eq:kl-reverse}.
Given a discretization $1 = t_S > t_{S-1} > \cdots > t_0 = 0$ with
step sizes $\Delta t_k = t_k - t_{k-1}$, the update is:
\begin{align}
  X_{t_{k-1}} &= X_{t_k} - \Delta t_k\!\Bigl[
    f(t_k)\,X_{t_k} \nonumber\\
    & + \frac{\gM^2(t_k)}{\sigM(t_k)}\,\veps_\theta(X_{t_k}, t_k)
    + \gM(t_k)\sum_{m=1}^{M} \psi_m(t_k)\,\bar{Z}_m
  \Bigr],\label{eq:kl-euler}
\end{align}
where we approximate the score via the $\veps$-parameterization:
$\nabla_X \log p_t^{(M)}(X_t) \approx -\veps_\theta(X_t, t)/\sigM(t)$
(here $\veps_\theta$ is trained at the prescribed level unless the KL loss of
\Cref{sec:kl-loss} is used; this gap is the mismatch analyzed in
\Cref{sec:mstar-prediction}).
The expansion coefficients $\{\bar{Z}_m\}$ are drawn once and reused at
every step, providing a \emph{structured noise trajectory} that
contrasts with the i.i.d.\ noise in the standard reverse SDE Euler.
KL-Euler uses all $S=1000$ time steps (the original DDPM schedule) and
serves as the baseline for the KL sampler family.

\subsubsection{KL-DPM-Solver\texorpdfstring{$(M, k)$}{(M, k)}}

The KL probability flow ODE admits the variation-of-constants (VoC)
formulation used by DPM-Solver~\cite{lu2022dpm}.  Substituting
$X_t = \alpha(t)\,u_t$ converts the ODE to
$\dot{u}_t = -\tfrac{1}{2}\gM^2(t)\,\veps_\theta / (\alpha(t)\,\sigM(t))$.
Integration from $s$ to $t$ gives
\begin{equation}\label{eq:kl-dpm-update}
  X_t = \frac{\alpha(t)}{\alpha(s)}\,X_s
  + \alpha(t)\!\int_{\lamM(s)}^{\lamM(t)}\!
    e^{-\lambda}\,\hat{\veps}_\theta\;\dd\lambda,
\end{equation}
where $\lamM(t) := \log\bigl(\sqrt{\abar_t}/\sigM(t)\bigr)$ is the
KL log-SNR.  The integral is approximated by Taylor expansion of
$\hat{\veps}_\theta$ in $\lambda$.

\textbf{Order 1.}
Freezing $\hat{\veps}_\theta$ at $\lambda_s$ yields the first-order
update, equivalent to KL-DDIM at $\eta=0$.

\textbf{Order 2 (multistep).}
Using two consecutive model evaluations $\hat{\veps}_s$, $\hat{\veps}_{s'}$
(with $s' < s$), the corrected prediction is
\[
  \hat{\veps}_{\mathrm{corr}} = \hat{\veps}_s
  + \frac{1}{2r}\bigl(\hat{\veps}_s - \hat{\veps}_{s'}\bigr),
  \quad r = \frac{\lamM(s) - \lamM(s')}{\lamM(t) - \lamM(s)}.
\]
The first-order update~\eqref{eq:kl-dpm-update} is then applied with
$\hat{\veps}_{\mathrm{corr}}$.
A key implementation detail is that the log-SNR schedule
$\lamM(t)$ differs from $\lambda(t)$, affecting the extrapolation ratio~$r$.
The ratio $\sigM(t)/\alpha(t)=\bigl(\sum_{m\le M}c_m(t)^2\bigr)^{1/2}$ is
strictly increasing in $t$ (verified numerically on our schedule for all
$M\in[2,1024]$), so $\lamM$ is strictly monotone and $r$ is well-defined at
every~$M$; $\sigM(t)$ itself is non-monotone in the high-noise region but
never enters the solver through a ratio.
In our experiments, order-2 with $M\!=\!256$ gives the best KL-DPM result.

\paragraph{KL-DPM-Solver++ and other samplers}
DPM-Solver++~\cite{lu2022dpmpp} expands the \emph{data prediction} $\hat{x}_0(\lambda)$ (order~2
is the standard guided sampler for Stable Diffusion). KL truncation enters only through the
data prediction, $\hat{x}_{0,\mathrm{KL}}=(x-\sigM\hat{\veps})/\alpha$, leaving all ODE
transition coefficients at the standard~$\sigma$ and preserving order-2 stability; on SD~v1.4
KL-DPM++ ($M\!=\!64$) improves over DPM-Solver++ and LoRA-KL-DPM++ ($M\!=\!32$) is best
(\Cref{tab:sd-fid}). More generally the $\sigma\to\sigM$ substitution extends to any
schedule-dependent sampler: for the learned predictor AMED~\cite{zhou2024amed}, KL-AMED replaces
only the step-size terms
$(\sigma_{\mathrm{next}}-\sigma_{\mathrm{cur}})\to(\sigM(\sigma_{\mathrm{next}})-\sigM(\sigma_{\mathrm{cur}}))$,
with predictor outputs and denoiser conditioning unchanged
(Supplementary App.~F).

\section{Experiments}
\label{sec:experiments}

\subsection{Setup}

\paragraph{Datasets.}
CIFAR-10 ($32\!\times\!32$, 50k training images),
CelebA~\cite{liu2015celeba} ($64\!\times\!64$, 202k images),
ImageNet~\cite{deng2009imagenet} ($64\!\times\!64$, 1.28M training images, 1000 classes),
and Stable Diffusion v1.4~\cite{rombach2022high} (COCO val2014, 512$\times$512).

\paragraph{Models and checkpoint abbreviations}
For CIFAR-10, we use two checkpoints: \textbf{PFDiff-790k}, the PFDiff
EMA pretrained checkpoint~\cite{wang2025pfdiff} used for sampler and
LoRA comparisons, and \textbf{Scratch-2040}, the UNet trained from
scratch for 2040 epochs to evaluate the KL training loss.
For CelebA, \textbf{CelebA-DDIM} denotes the pretrained DDIM checkpoint
from~\cite{songDenoisingDiffusionImplicit2021}.
For ImageNet, \textbf{EDM-ImageNet64} denotes the pretrained
EDM checkpoint~\cite{karrasElucidatingDesignSpace2022} with the VE-SDE
parameterization ($\sigma_{\min}\!=\!0.002$, $\sigma_{\max}\!=\!80$,
polynomial schedule $\rho\!=\!7$).
The \textbf{AMED-predictor}~\cite{zhou2024amed} is a lightweight network
trained on top of EDM-ImageNet64 to predict per-step adaptive parameters
(4 steps with AFS, NFE$\!=\!5$).
For Stable Diffusion, \textbf{SD-v1.4} denotes the public Stable
Diffusion v1.4 checkpoint with classifier-free guidance ($w\!=\!7.5$).
For orientation across checkpoints: KL-DDIM ($M{=}128$, $S{=}20$, $\eta{=}0$)
attains FID-50k $5.07$ on PFDiff-790k, $4.77$ on the epoch-1080 scratch
checkpoint, and $4.98$ on Scratch-2040; throughout, every quoted FID names its
checkpoint.

\paragraph{Evaluation}
FID~\cite{heusel2017gans} computed with 50k generated samples (FID-50k)
for main results and 5k samples for ablations.
All experiments use quadratic timestep scheduling; all samplers, baselines
included, share this grid, so every comparison isolates the noise magnitude at
fixed discretization.
Generation uses 4 GPUs with per-GPU seed offsets, and our conclusions rest on
the shape of entire $M$- and $S$-sweeps rather than on isolated pairwise
differences. Code, configurations, and evaluation scripts will be released upon
publication.

\subsection{Train-from-Scratch: Locating the Effective Mode Count \texorpdfstring{$M^*$}{M*}}
\label{sec:train-scratch}

The train-from-scratch setting is the \emph{controlled} experiment for the
prediction of \Cref{sec:mstar-prediction}: unlike a fixed pretrained
checkpoint, here we can \emph{set} the training order~$M'$ and watch the
network's own sweet spot~$M^*$ respond. We answer the two questions posed in
\Cref{sec:mstar-prediction} on CIFAR-10 (Scratch-2040; protocol in
\Cref{sec:experiments}), and show that the sampling-side and training-side
sweeps are two views of the same effective mode count~$M^*$.

\begin{figure}[t]
\centering
\includegraphics[width=0.62\columnwidth]{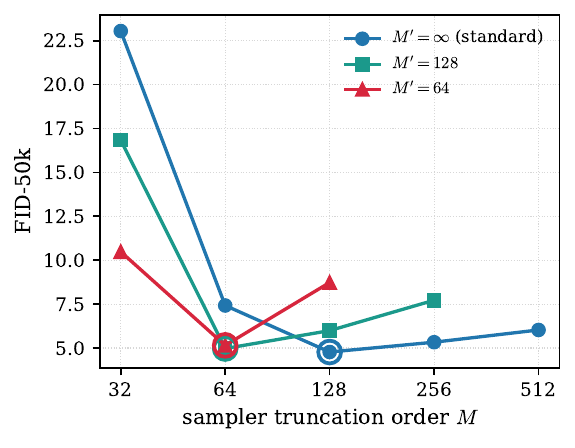}
\caption{CIFAR-10 epoch-1080 KL-DDIM $M$-sweep (FID-50k, $S\!=\!20$,
$\eta\!=\!0$, quadratic skip; log-scale~$M$). Each curve is one
train-from-scratch checkpoint with training order~$M'$; the highlighted point
is each curve's sweet spot~$M^*$. Reducing~$M'$ from~$\infty$ (standard) to
$128$ to $64$ moves the sweet spot left ($M^*\!=\!128\!\to\!64\!\to\!64$) until
$M^*\!=\!M'$, the fixed point of the descending-$M^*$ procedure.}
\label{fig:mstar-msweep}
\end{figure}

\paragraph{Sampling axis: $M^*$ on a standard checkpoint.}
Take the standard train-from-scratch checkpoint ($M'\!=\!\infty$, epoch 1080)
and, without any retraining, sweep the KL-DDIM order~$M$
($\eta\!=\!0$, $S\!=\!20$).
The FID curve is strongly non-monotone (\Cref{fig:mstar-msweep}): it falls
from $23.03$ at $M\!=\!32$ to a minimum of $4.77$ at $M\!=\!128$, then rises
monotonically ($5.29$ at $M\!=\!256$, $6.00$ at $M\!=\!512$) as
$\sigM\!\to\!\sigma$. The standard DDIM sampler (the $M\!=\!\infty$ limit) gives
$6.63$ on this checkpoint, so the sweet spot at $M^*\!=\!128$ is a $28\%$
training-free improvement. A network trained to resolve the infinite-mode
driver thus behaves, at sampling time, as if its effective mode count were
finite, matching the effective-noise prediction of \Cref{prop:sweetspot}.

\begin{figure}[t]
\centering
\includegraphics[width=0.72\columnwidth]{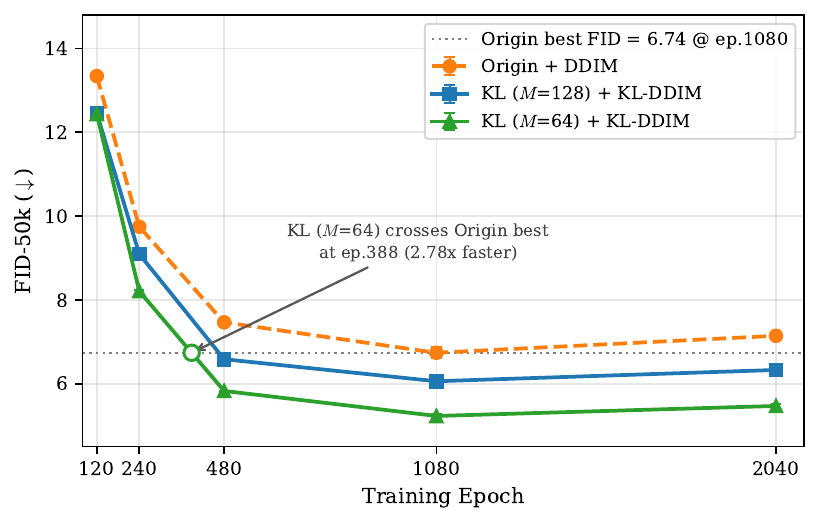}
\caption{CIFAR-10 Scratch-2040 training convergence (matched $M\!=\!M'$). The
dotted line marks the best Origin + DDIM checkpoint (FID\,=\,6.74 at epoch
1080), which KL($M'\!=\!64$) surpasses already at epoch 480; interpolation
between epochs 240 and 480 places the crossing at epoch $\approx 388$
($\approx2.8\times$ speedup).}
\label{fig:training-convergence}
\end{figure}

\paragraph{Training axis: matched KL training.}
Training instead under the KL forward process at a finite order~$M'$ and
evaluating with the matched KL-DDIM ($M\!=\!M'$) improves the diagonal
$M\!=\!M'$ result monotonically as~$M'$ decreases toward~$M^*$
(\Cref{tab:train-scratch}, \Cref{fig:training-convergence}).
KL($M'\!=\!64$)+KL-DDIM reaches FID\,=\,5.47 at epoch 2040 versus 7.14 for
Origin+DDIM, while KL($M'\!=\!128$) reaches 6.33.
KL($M'\!=\!64$) surpasses the \emph{best} Origin+DDIM checkpoint at any epoch
(FID\,=\,6.74, epoch 1080) already at epoch 480 (FID\,=\,5.83), with
$2.25\times$ fewer epochs; linear interpolation between epochs 240 and 480
places the crossing at epoch~$\approx 388$, a $\approx2.8\times$
training-convergence speedup under matched 20-step evaluation (highlighted
line in \Cref{fig:training-convergence}). Best checkpoint against best
checkpoint, the comparison is 6.74 vs.\ 5.23. Each entry of
\Cref{tab:train-scratch} is a mean over three sampling seeds; the spread is
small ($\le\!0.13$ FID on the noisiest row), and we report it here as a
representative example rather than for every table.

\paragraph{The two axes meet: a descending-$M^*$ procedure.}
The sampling-side and training-side views coincide. Per-checkpoint
$M$-sweeps (\Cref{fig:mstar-msweep}) show that the sweet spot~$M^*$ of a
KL-trained network sits \emph{at or below} its training order~$M'$, and that
each retraining at $M'\!=\!M^*$ shifts the whole curve down and to the left:
\begin{itemize}[leftmargin=1.4em,itemsep=1pt,topsep=2pt]
  \item $M'\!=\!\infty$ (standard): $M$-sweep minimum at $M^*\!=\!128$;
  \item $M'\!=\!128$: minimum moves to $M^*\!=\!64$ ($M\!=\!64$ beats
  $M\!=\!128$, $4.97$ vs $5.99$), i.e.\ the network still cannot use all $128$
  trained modes;
  \item $M'\!=\!64$: minimum is at $M^*\!=\!64\!=\!M'$ ($M\!=\!64$ beats both
  $M\!=\!32$ and $M\!=\!128$), so the gap $|M^*\!-\!M'|$ has closed and the
  procedure terminates.
\end{itemize}
This yields a simple recipe: train at $M'$, locate $M^*$ by a sampling-side
$M$-sweep, retrain at $M'\!\leftarrow\!M^*$, and repeat until $M^*\!=\!M'$.
The FID gain from the search itself is modest; its value is
mechanistic---it exhibits a checkpoint at the fixed point $M^*\!=\!M'$, confirming
$M^*$ as a real, controllable property recovered by either probe. We present it as an
empirical heuristic: it terminates in two iterations here, but we do not
establish general convergence.

\paragraph{Interpretation.}
The KL loss differs from the standard loss only in the reduced level
$\sigma_{M'}(t)$ and recovers it as $M'\!\to\!\infty$ (\Cref{sec:kl-loss}), so
the framework extends to training. The descending procedure terminates where
the sampling-side sweep confirms the network realizes the level it was trained
at; this also
explains why $M'\!=\!64$ beats $M'\!=\!128$ even though the latter is the
closer approximation to the full SDE by \Cref{thm:convergence}---the
$M'\!=\!128$ checkpoint's own sweet spot sits at $M^*\!=\!64$. Faster
convergence at reduced training levels is consistent with known effects of the
training noise-level distribution~\cite{hangEfficientDiffusionTraining2023}
and is not by itself evidence for the mechanism; the identifying evidence is
the sweep structure, an inference-time measurement. The study is a
matched-budget comparison (shared architecture, optimizer, and epochs), so its
absolute FIDs are not comparable to the separately trained PFDiff-790k.

Having established~$M^*$ as a real, controllable property of the trained
network, we next show that the \emph{same}~$M^*$ can be read off \emph{fixed}
pretrained checkpoints, where it yields training-free gains across datasets,
samplers, and scales (\Cref{sec:pretrained-samplers}).

\begin{table}[t]
\centering
\caption{CIFAR-10 Scratch-2040 matched train-from-scratch comparison
($M\!=\!M'$). FID-50k with $S\!=\!20$, $\eta\!=\!0$, quadratic skip;
mean\,$\pm$\,std over three sampling seeds.}
\label{tab:train-scratch}
\setlength{\tabcolsep}{4pt}\renewcommand{\arraystretch}{1.05}\small
\begin{tabular}{lccc}
\toprule
Epoch & Origin + DDIM & KL($M'$=128) & KL($M'$=64) \\
\midrule
120  & $13.34\pm0.08$ & $12.45\pm0.08$ & $12.44\pm0.08$ \\
240  & $9.75\pm0.07$  & $9.09\pm0.06$  & $\best{8.21}\pm0.04$ \\
480  & $7.47\pm0.07$  & $6.58\pm0.05$  & $\best{5.83}\pm0.01$ \\
1080 & $6.74\pm0.13$  & $6.06\pm0.09$  & $\best{5.23}\pm0.05$ \\
2040 & $7.14\pm0.10$  & $6.33\pm0.08$  & $\best{5.47}\pm0.05$ \\
\bottomrule
\end{tabular}
\end{table}

\subsection{Pretrained Model: KL Samplers}
\label{sec:pretrained-samplers}

Using PFDiff-790k (no retraining), we evaluate KL
samplers against standard baselines.

\paragraph{KL-Euler}
As the simplest KL sampler, KL-Euler uses all $S\!=\!1000$ DDPM steps.
At $M\!=\!128$ it achieves FID\,=\,7.83 (FID-5k), within 0.06 of DDPM
(7.77) and clearly ahead of DDIM-100 (8.91; all FID-5k), showing that the
structured KL noise trajectory costs essentially nothing even for the basic
1000-step solver
(full $M$-sweep in Supplementary App.~D).

\paragraph{KL-DDIM}
KL-DDIM ($M\!=\!128$) is the best deterministic sampler at every step count
(\Cref{tab:main-det}), cutting FID by 27\% at $S\!=\!20$ ($6.90\!\to\!5.07$) and
staying ahead through $S\!=\!100$ ($4.19\!\to\!3.64$).

\paragraph{KL-DPM-Solver}
KL-DPM O2 ($M\!=\!256$) improves on DPM-Solver O2 at every step count
(\Cref{tab:main-det}), e.g.\ FID\,=\,4.11 vs.\ 4.61 at $S\!=\!20$ (an 11\%
improvement). At very aggressive step counts the substitution breaks down
($S\!=\!4$: 72.1 vs.\ 52.1, Supplementary App.~E),
delimiting the few-step validity of the KL schedule in high-order solvers.

\paragraph{Cross-sampler analysis}
Two patterns emerge across the KL samplers.
First, the benefit grows with solver order (Euler $\to$ DDIM $\to$ DPM O2):
KL-Euler only matches its baseline (7.83 vs.\ 7.77 FID-5k, a 0.8\%
regression), while KL-DDIM improves 27\% over DDIM at $S\!=\!20$ and
KL-DPM O2 improves 11\% over DPM-Solver O2.
This extends further to learned-predictor samplers
(\Cref{sec:generality}).
Second, the optimal $M$ increases with solver quality:
$M\!=\!128$ for DDIM, $M\!=\!256$ for DPM O2, and $M\!=\!950$ for AMED
on ImageNet (\Cref{fig:msweep-universal}).
Measured at the start level, this variation is small:
$\sigM(t_S)/\sigma(t_S)\approx0.994$ at $M\!=\!128$ versus $\approx0.997$ at
$M\!=\!256$, a sub-percent difference. Because $\sigM(t_S)$ saturates as $M$
grows, the integer $M$ is a stretched coordinate that magnifies small level
differences: the operating level identified by the sweeps is nearly
sampler-invariant even though $M^*$ is not; why the swept order depends on the
sampler remains open (\Cref{sec:conclusion}).

\paragraph{Effective noise schedule of pretrained models.}
KL samplers improve checkpoints never trained at the KL noise level, contrary to the usual
assumption that training and sampling schedules must
match~\cite{linCommonDiffusionNoise2024}. The $M$-sweep is non-monotone across all datasets,
samplers, and SDE frameworks (\Cref{fig:msweep-universal}): very small~$M$ underperforms (the
gap $|\sigma-\sigM|$ too large), large~$M$ approaches the baseline as $\sigM\!\to\!\sigma$
(e.g.\ KL-DDIM at $M\!=\!512$ gives FID\,=\,6.27 vs.\ 6.90), and an intermediate~$M$ is
best---the steep collapse arm, interior minimum near $\sigM\!\approx\!\seff$, and gentle
over-dispersion arm of \Cref{prop:sweetspot,fig:sweetspot}. A monotone mismatch would give no
interior minimum; the interior optimum signals that the network's effective noise level lies below
$\sigma(t)$, as expected a priori from finite-sample shrinkage (\Cref{def:eff}). This
sweet-spot order is the effective mode count~$M^*$; \Cref{sec:train-scratch} shows the
training-side sweep recovers the same~$M^*$.

\paragraph{Distinguishing ODE conditioning from schedule matching}
One alternative is that $\sigM<\sigma$ merely better-conditions the probability-flow ODE.
Three observations disfavor it: (i)~stochastic KL-DDIM ($\eta>0$, Supplementary
Table~IV) gains \emph{more} than the deterministic case, whereas
conditioning benefits should shrink under injected noise; (ii)~the gain persists at large $S$
(\Cref{tab:main-det}, $S\!=\!100$), where discretization error is negligible; and
(iii)~KL-DAPS (\Cref{tab:daps-phase-retrieval}) improves an annealing sampler that integrates
no probability-flow ODE. These favor the schedule-matching interpretation.

\begin{table}[t]
\centering
\caption{Deterministic samplers on CIFAR-10 PFDiff-790k, FID-50k,
quadratic skip.}
\label{tab:main-det}
\setlength{\tabcolsep}{4pt}\renewcommand{\arraystretch}{0.95}\small
\begin{tabular}{llrrrr}
\toprule
Sampler & $M$ & $S\!=\!10$ & $S\!=\!20$ & $S\!=\!50$ & $S\!=\!100$ \\
\midrule
DDIM & --- & 13.51 & 6.90 & 4.69 & 4.19 \\
DPM-Solver O2 & --- & 6.60 & 4.61 & 4.34 & 4.28 \\
\midrule
KL-DDIM & 128 & 9.45 & 5.07 & \best{3.84} & \best{3.64} \\
KL-DPM O2 & 128 & 6.21 & 4.54 & 4.28 & 4.20 \\
KL-DPM O2 & 256 & \best{5.76} & \best{4.11} & 3.91 & 3.85 \\
\bottomrule
\end{tabular}
\end{table}

Stochastic KL-DDIM ($\eta > 0$) also outperforms DDIM at every matched
$\eta$, with the largest gain at $\eta\!=\!0.5$ ($-36\%$); full
KL-DPM $M$/$S$-sweep curves and the stochastic table appear in
Supplementary App.~E.

\subsection{Why KL? A Uniform-Rescaling (\texorpdfstring{$\nu$}{nu}-Sampler) Control}
\label{sec:rho}
The analysis attributes the sweet spot to the effective-noise mismatch, not to the KL basis:
\Cref{prop:sweetspot-exact} assumes only that the family is pointwise monotone in~$M$, so it
predicts the same U-shape for any monotone reduced schedule and is \emph{silent} on which
family reaches the lower optimum. Whether KL beats a flat rescale is therefore an empirical
question.
We test the simplest constant-$\nu$ family, the \textbf{$\nu$-sampler}
$\sigma^\dagger_M(t)=\nu_M\,\sigma(t)$ with $\nu_M=1-1/M$ ($\nu$-DDIM), the schedule-space
analogue of the constant $\varepsilon$-scaling of Ning et al.~\cite{ningExposureBiasElucidating2024}.
On CIFAR-10 ($S\!=\!20$, $\eta\!=\!0$), $\nu$-DDIM indeed produces the same interior
optimum (\Cref{tab:rho}), confirming the sweet spot is a property of the mismatch, not the
basis. Yet the KL optimum is better: $4.98$ at $M^*\!=\!128$ beats the $\nu$-sampler
optimum ($5.37$ at $M^*\!\approx\!192$--$224$) at a \emph{stronger}
modification; away from their optima the two families cross (for
$M\!\gtrsim\!224$ the $\nu$ row is lower), so
the comparison of interest is between the family optima. We attribute the
remaining gain to KL's $t$-dependent contraction---the interior-profile effect
the analysis is silent on---as the natural interpretation; the
control compares the two families' optima and does not isolate the mechanism
itself. We therefore use KL schedules throughout. This control is a
single-dataset, single-sampler comparison (CIFAR-10, KL-DDIM, $S\!=\!20$), and
since we compare the \emph{optima} of the two families, it does not depend on
the particular indexing $\nu_M\!=\!1-1/M$ of constant rescalings.

\begin{table}[t]
\centering
\caption{$\nu$-DDIM (uniform rescaling, $\nu_M\!=\!1\!-\!1/M$) vs.\ KL-DDIM on CIFAR-10
(Scratch-2040, FID-50k, $S\!=\!20$, $\eta\!=\!0$). Sweet points:
KL $M^*\!=\!128$ ($4.98$); $\nu$-sampler $M^*\!\approx\!192$--$224$ ($5.37$).}
\label{tab:rho}
\setlength{\tabcolsep}{4pt}\renewcommand{\arraystretch}{0.95}\small
\begin{tabular}{lrrrrrr}
\toprule
$M$ & 64 & 128 & 160 & 192 & 224 & 256 \\
\midrule
$\nu$-DDIM & 14.90 & 6.12 & 5.54 & \best{5.37} & \best{5.37} & 5.42 \\
KL-DDIM & 7.17 & \best{4.98} & 5.13 & 5.33 & 5.49 & 5.65 \\
\bottomrule
\end{tabular}
\end{table}

\subsection{Generality: Datasets, Samplers, and Scale}
\label{sec:generality}

\begin{figure*}[t]
\centering
\includegraphics[width=0.94\textwidth]{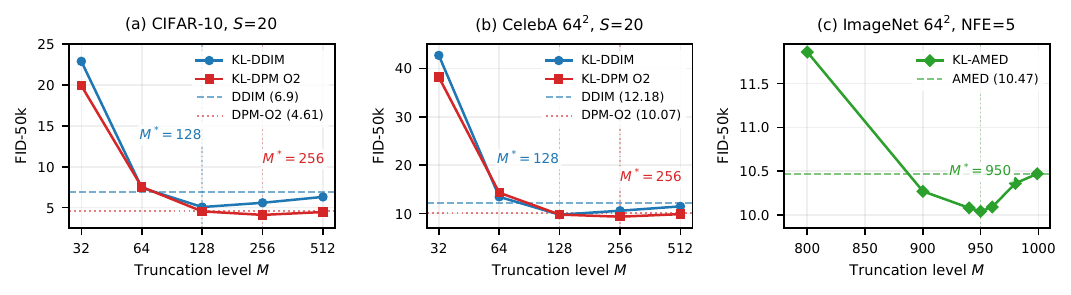}
\caption{$M$-sweep FID curves across datasets and samplers. Dashed lines
mark baselines ($M\!\to\!\infty$). All panels share the same non-monotone
shape: the sweet spot (arrows) beats the baseline, then FID degrades
for small~$M$. The optimal~$M$ differs by sampler and dataset.
(a)~CIFAR-10 PFDiff-790k, $S\!=\!20$, FID-50k.
(b)~CelebA~$64^2$, $S\!=\!20$, FID-50k.
(c)~ImageNet~$64^2$ EDM, NFE$\!=\!5$, FID-50k.}
\label{fig:msweep-universal}
\end{figure*}

\Cref{fig:msweep-universal} collects $M$-sweep curves across three
datasets and sampler families. All panels exhibit the same non-monotone
pattern identified on CIFAR-10 in \Cref{sec:pretrained-samplers}:
FID improves as $M$ decreases from the baseline, reaches a sweet spot,
then degrades for small~$M$.
The optimal~$M$ varies by configuration, but the shape is invariant.
\Cref{tab:kl-summary} summarizes the best result per configuration.

\begin{table}[t]
\centering
\caption{Summary of best KL sampler improvements at the sweet-spot~$M$.}
\label{tab:kl-summary}
\setlength{\tabcolsep}{4pt}\renewcommand{\arraystretch}{0.95}\small
\footnotesize
\begin{tabular}{llcrrc}
\toprule
Dataset & Sampler & $S$ & Baseline & Best $M$ & KL FID \\
\midrule
CIFAR-10   & KL-DDIM   & 20 & 6.90  & 128 & \best{5.07} \\
CIFAR-10   & KL-DPM O2 & 20 & 4.61  & 256 & \best{4.11} \\
CelebA     & KL-DDIM   & 20 & 12.18 & 128 & \best{9.76} \\
CelebA     & KL-DPM O2 & 20 & 10.07 & 256 & \best{9.35} \\
ImageNet   & KL-AMED   & 5  & 10.47 & 950 & \best{10.04} \\
\bottomrule
\end{tabular}
\end{table}

\paragraph{CelebA \texorpdfstring{$64\!\times\!64$}{64x64}}
\label{sec:celeba}
\Cref{tab:celeba-det} shows consistent improvements: KL-DDIM
($M\!=\!128$) reduces FID by 20\% at $S\!=\!20$ (12.18 $\to$ 9.76)
and 10\% at $S\!=\!100$ (9.71 $\to$ 8.77).
The $M$-sweep pattern (\Cref{fig:msweep-universal}b) matches CIFAR-10,
with minimum near $M\!=\!128$ for KL-DDIM and $M\!=\!256$ for KL-DPM.
The optimal orders are unchanged from CIFAR-10 despite the fourfold change in
resolution and dataset.

\begin{table}[t]
\centering
\caption{Deterministic samplers on CelebA-DDIM ($64\!\times\!64$),
FID-50k.}
\label{tab:celeba-det}
\setlength{\tabcolsep}{4pt}\renewcommand{\arraystretch}{0.95}\small
\begin{tabular}{llrrrr}
\toprule
Sampler & $M$ & $S\!=\!10$ & $S\!=\!20$ & $S\!=\!50$ & $S\!=\!100$ \\
\midrule
DDIM & --- & 17.46 & 12.18 & 10.17 & 9.71 \\
DPM-O2 & --- & 12.67 & 10.07 & 9.65 & 9.58 \\
\midrule
KL-DDIM & 128 & 13.54 & \best{9.76} & \best{8.75} & \best{8.77} \\
KL-DPM O2 & 256 & \best{11.26} & 9.35 & 9.12 & 9.04 \\
\bottomrule
\end{tabular}
\end{table}

\paragraph{KL-AMED on ImageNet \texorpdfstring{$64\!\times\!64$}{64x64}}
\label{sec:kl-amed}
AMED~\cite{zhou2024amed} represents a fundamentally different sampler
class: a trained predictor network outputs per-step adaptive parameters
$(r, s_{\mathrm{dir}}, s_{\mathrm{time}})$ from denoiser encoder features,
targeting aggressive NFE$\!=\!5$.
KL-AMED applies the same $\sigma \to \sigM$ step-size substitution
(Supplementary App.~F); predictor outputs and denoiser conditioning are
unchanged, and the predictor is reused frozen from the released checkpoint.
The VE-grid construction of $\sigM$ is a discrete analogue outside the VP
setting of \Cref{thm:convergence} and is justified here empirically.
\Cref{fig:msweep-universal}c shows the $M$-sweep on
ImageNet~$64\!\times\!64$ (EDM, VE-SDE):
the sweet spot is $M\!=\!950$ (FID\,=\,10.04 vs baseline 10.47,
$-4.1\%$), with a broad flat optimum $M \in [940, 960]$.
(Baseline FIDs are from our reproduction using the released checkpoints
and may differ slightly from the values in the original papers.)
Despite operating in a different SDE framework (VE vs VP),
on a larger class-conditional dataset (1000 classes, 1.28M images),
and with a learned rather than formula-based sampler,
the non-monotone sweet-spot pattern is preserved.

\paragraph{Stable Diffusion v1.4}
\label{sec:sd}
To demonstrate applicability at scale, we evaluate KL samplers on Stable
Diffusion v1.4 in the $4\!\times\!64\!\times\!64$ latent space
(\Cref{tab:sd-fid}).
KL-DDIM ($M\!=\!64$) improves over DDIM at both NFE values
(18.10 vs 23.23 at NFE$\!=\!5$; 15.15 vs 17.42 at NFE$\!=\!10$).
KL-DPM++ ($M\!=\!64$) improves over DPM-Solver++ at NFE$\!=\!5$
(16.24 vs 18.31) but is slightly worse at NFE$\!=\!10$
(14.74 vs 14.52), suggesting that the KL schedule benefit is
strongest in the few-step regime where discretization error dominates.
LoRA-KL-DPM++ ($M\!=\!32$) achieves the best NFE$\!=\!5$ FID
(13.38), while LoRA-KL-DDIM ($M\!=\!32$) achieves the best
NFE$\!=\!10$ FID (12.86).
The lower optimal $M\!=\!64$ (vs.\ $M\!=\!128$ for CIFAR-10/CelebA)
may reflect the reduced temporal complexity of latent-space diffusion.

\begin{table}[t]
\centering
\caption{FID-10k on SD-v1.4 (COCO val2014,
$w\!=\!7.5$). DPM-Solver++ is the order-2 multistep guided sampler
from~\cite{lu2022dpmpp}; KL-DPM++ applies KL data prediction
on top of DPM-Solver++ (see text).}
\label{tab:sd-fid}
\setlength{\tabcolsep}{4pt}\renewcommand{\arraystretch}{0.95}\small
\small
\begin{tabular}{lccc}
\toprule
Sampler & $M$ & NFE$=5$ & NFE$=10$ \\
\midrule
DDIM                                   & ---  & 23.23 & 17.42 \\
DPM-Solver++~\cite{lu2022dpmpp}        & ---  & 18.31 & 14.52 \\
KL-DDIM                                & 64   & 18.10 & 15.15 \\
KL-DPM++                               & 64   & 16.24 & 14.74 \\
LoRA-KL-DDIM                           & 32   & 14.24 & \best{12.86} \\
LoRA-KL-DPM++                          & 32   & \best{13.38} & 13.74 \\
\bottomrule
\end{tabular}
\end{table}

\begin{figure}[t]
\centering
\includegraphics[width=0.92\columnwidth]{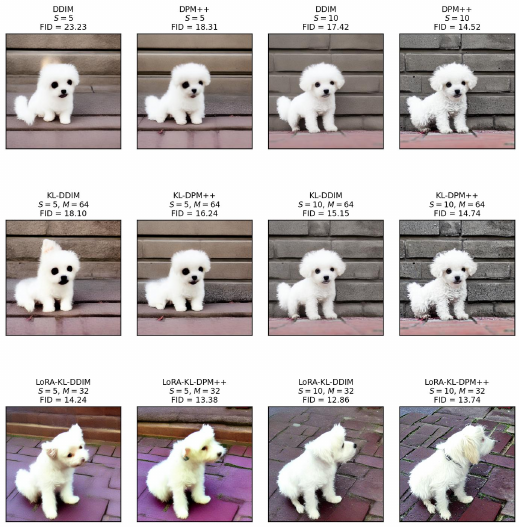}
\caption{SD-v1.4 visual comparison on COCO val2014.
Prompt: ``\emph{small white dog by green post on brick sidewalk}.''
Row~1: DDIM and DPM-Solver++ baselines.
Row~2: KL-DDIM ($M\!=\!64$) and KL-DPM++ ($M\!=\!64$).
Row~3: LoRA-KL-DDIM ($M\!=\!32$) and LoRA-KL-DPM++ ($M\!=\!32$).
All panels share the same initial noise.}
\label{fig:sd-table6-visual}
\end{figure}

\paragraph{Synthesis}
Across six sampler--dataset combinations
(\Cref{tab:kl-summary,tab:sd-fid,fig:msweep-universal}) the non-monotone
sweet-spot pattern is invariant, while the optimal order varies (e.g.\ $M\!=\!128$
for KL-DDIM, $256$ for KL-DPM O2, $950$ for KL-AMED, $64$ for latent-space SD).
Every tested checkpoint, regardless of architecture, dataset, or SDE framework,
thus behaves as if internalizing a noise level slightly below $\sigma(t)$ that
the KL family $\{\sigM(t)\}$ probes; the sole regression among the pairs tested
in these tables is KL-DPM++ at NFE$\,=\,$10 on SD (\Cref{tab:sd-fid});
KL-Euler is on par with DDPM, and the few-step breakdown at $S\!=\!4$ is
delimited in Supplementary App.~E. In practice the sweep
need not use the full protocol: on CIFAR-10, where we computed both, a cheap
FID-5k sweep selects the same sweet-spot order as FID-50k (KL-DDIM $M\!=\!128$;
KL-DPM O2 $M\!=\!256$ at $S\!=\!10$ and $S\!=\!20$), so the full evaluation
budget is spent once at the selected order.

\subsection{LoRA Fine-Tuning}
\label{sec:lora}

When $M$ is small, the noise gap $|\sigma(t) - \sigM(t)|$ is large and
the pretrained denoiser is mismatched.
LoRA~\cite{hu2022lora} with rank $r\!=\!4$ efficiently adapts the model
to the KL noise schedule.
This experiment intentionally uses PFDiff-790k: the goal is to test whether KL-Diffusion can benefit from
existing pretrained models rather than requiring training from scratch.
\Cref{fig:lora-comparison} (raw values in Supplementary Table~VI) shows that LoRA
dramatically improves FID for $M \leq 64$ (up to 77\% at $M\!=\!10$)
but hurts for $M \geq 128$ where the pretrained model is
already well-matched, with crossover between $M\!=\!64$ and $M\!=\!128$.

\paragraph{Adaptation analysis}
The crossover has a clean interpretation. For large $M$, $\sigM\!\approx\!\sigma$
(\Cref{thm:convergence}) and the pretrained denoiser is already well-calibrated, so LoRA's
extra parameters add a residual bias (FID floors near $5.1$ vs.\ $3.6$ unmodified). For small
$M$ the gap $|\sigma-\sigM|$ is large ($0.048$ at $M\!=\!20$, $0.091$ at $M\!=\!10$) and the
denoiser overestimates noise; LoRA learns a low-rank bias toward the $\sigM$ scale, giving up
to $4.4\times$ improvement at $M\!=\!10$. That rank $r\!=\!4$ ($<\!0.1\%$ parameters) suffices
confirms a low-dimensional correction, consistent with the spectral shift having rank $\leq M$.

\begin{figure}[t]
\centering
\includegraphics[width=0.72\columnwidth]{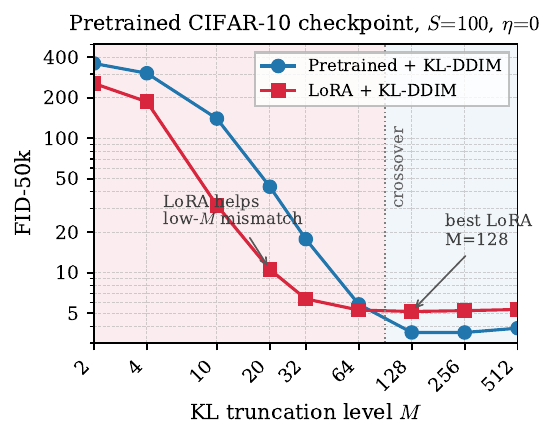}
\caption{LoRA adaptation on CIFAR-10 PFDiff-790k (FID-50k). Both curves use
KL-DDIM with $S\!=\!100$ and $\eta\!=\!0$.
LoRA improves low-$M$ settings where the KL noise schedule is most
mismatched to the pretrained denoiser; the curves cross between
$M\!=\!64$ and $M\!=\!128$, beyond which the unmodified pretrained
model is superior.}
\label{fig:lora-comparison}
\end{figure}

\subsection{Additional Support: Inverse Problem Solving}
\label{sec:daps-phase}

To test whether the KL truncation remains useful outside unconditional
generation, we plug the same KL noise schedule into Decoupled Annealing
Posterior Sampling (DAPS)~\cite{zhangDAPS2025} and evaluate the nonlinear
FFHQ-$256$ phase-retrieval task. Given a clean image $x_0$ and a noisy
measurement $y$, the inverse problem is
\begin{equation}
  y = \mathcal{A}(x_0) + n,
  \qquad n \sim \mathcal{N}(0,\beta_y^2 I),
\end{equation}
with the oversampled Fourier-amplitude operator
$\mathcal{A}(x) = |\mathcal{F}(\operatorname{Pad}(x))|$, oversampling ratio
$2.0$ and $\beta_y = 0.05$, matching the DAPS paper. Posterior sampling targets
$p(x_0 \mid y) \propto p(y \mid x_0)\,p_{\theta}(x_0)$.
KL-DAPS keeps the DAPS reconstruction pipeline unchanged and only replaces
the Brownian marginal scale by our KL scale from~\eqref{eq:sigM}, namely
$x_t^{(M)} = \alpha(t)\,x_0 + \sigma_M(t)\,\epsilon$ with
$\epsilon \sim \mathcal{N}(0,I)$,
so the annealing schedule now follows the $M$-truncated variance
$\sigma_M(t)^2$ instead of the original $\sigma(t)^2$.

\begin{figure*}[t]
\centering
\includegraphics[width=0.92\textwidth]{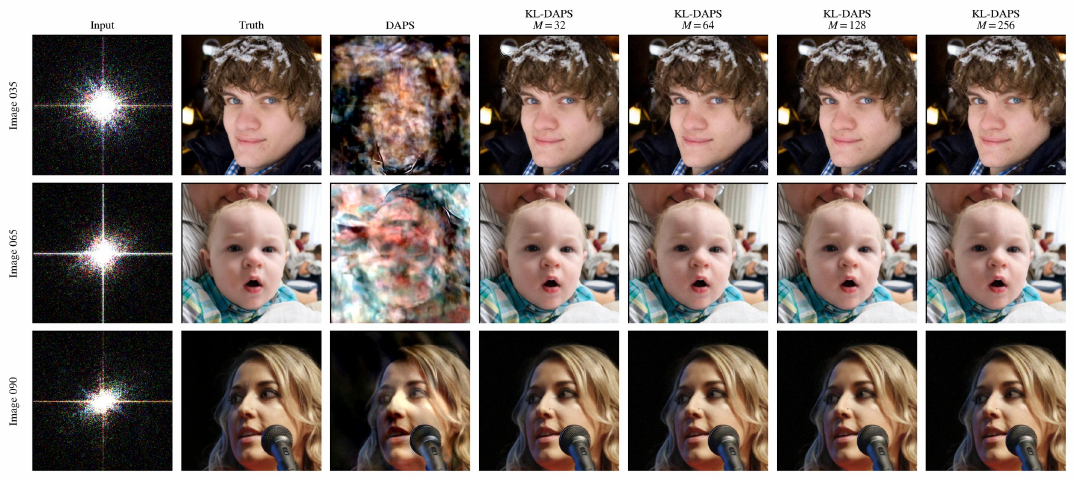}
\caption{FFHQ-$256$ phase retrieval visual comparison. Columns show the
measurement input, ground truth, reproduced DAPS reconstruction, and KL-DAPS
reconstructions for $M \in \{32,64,128,256\}$. For each method, the displayed
reconstruction is the highest-PSNR sample among the four runs used in the
paper-style best-of-four protocol. The three rows correspond to the validation
images with the largest per-image PSNR gains of KL-DAPS($M=32$) over DAPS among
cases where KL-DAPS also improves SSIM and LPIPS, namely indices 35, 65, and
90.}
\label{fig:daps-phase-visual}
\end{figure*}

We use the 100-image FFHQ-$256$ validation subset and the paper-style
\emph{best-of-four} protocol for phase retrieval, applied identically to DAPS
and KL-DAPS. As shown in
\Cref{tab:daps-phase-retrieval}, KL-DAPS improves the reproduced DAPS
baseline for all tested $M \in \{32,64,128,256\}$, with $M=32$ best overall;
the differences across $M$ are small. This setting improves PSNR
from $29.95$ to $30.68$ (+0.73\,dB), SSIM from $0.774$ to $0.815$, and LPIPS
from $0.178$ to $0.152$. The visual comparison in
\Cref{fig:daps-phase-visual} shows the same trend: KL-DAPS recovers cleaner
facial structure than the reproduced DAPS baseline across representative
examples, with $M=32$ typically the sharpest setting. This additional
inverse-problem experiment supports
the same conclusion as the generative benchmarks: a finite intermediate KL
truncation can provide a better effective noise schedule than the original
Brownian process even when the diffusion prior is used inside a posterior
sampler.

\begin{table}[t]
\centering
\caption{FFHQ-$256$ phase retrieval with DAPS and KL-DAPS. Means over 100
validation images under the DAPS paper-style best-of-four protocol.
The DAPS baseline is our reproduction using the released checkpoint.}
\label{tab:daps-phase-retrieval}
\setlength{\tabcolsep}{4pt}\renewcommand{\arraystretch}{0.95}\small
\small
\begin{tabular}{lrrrr}
\toprule
Method & $M$ & PSNR$\uparrow$ & SSIM$\uparrow$ & LPIPS$\downarrow$ \\
\midrule
DAPS~\cite{zhangDAPS2025} & --- & 29.95 & 0.774 & 0.178 \\
KL-DAPS & 32  & \best{30.68} & \best{0.815} & \best{0.152} \\
KL-DAPS & 64  & 30.56 & 0.800 & 0.159 \\
KL-DAPS & 128 & 30.67 & 0.799 & 0.159 \\
KL-DAPS & 256 & 30.44 & 0.791 & 0.165 \\
\bottomrule
\end{tabular}
\end{table}

\subsection{Alternative Expansion: L\'evy--Ciesielski (LC)}
\label{sec:lc-expansion}

The KL expansion uses cosine eigenfunctions $\{\psi_m\}$ because they
are \emph{optimal} in the $L^2$ sense---they minimize the mean-square
truncation error for any given $M$
(\Cref{sec:parseval})~\cite{pavliotisStochasticProcessesApplications2014}.
A natural question is whether other complete orthonormal bases yield
comparable results.

\paragraph{LC construction}
The L\'evy--Ciesielski (LC) construction represents Brownian motion
using the Haar wavelet (Schauder) basis.
An $N$-level LC expansion uses $M_{\mathrm{eff}} = 2^{N+1}$ basis
functions organized by dyadic resolution levels.
At level $N\!=\!7$ ($M_{\mathrm{eff}}\!=\!256$), the LC expansion gives
a marginal variance $\sigma_{\mathrm{LC}}^{(N)}(t)$ that converges to
$\sigma(t)$ in a piecewise-linear fashion.
The resulting LC samplers (LC-Euler, LC-DDIM, LC-DPM) are defined
analogously to their KL counterparts, replacing $\sigM(t)$ with
$\sigma_{\mathrm{LC}}^{(N)}(t)$; the explicit LC coefficients and sampler
updates are given in Supplementary App.~A.

\paragraph{Results}
\Cref{tab:lc-comparison} compares KL and LC samplers.
The LC variants consistently improve over the DDIM/DPM baselines,
confirming that \emph{finite-dimensional noise truncation} itself is
beneficial.
The better basis, however, depends on the solver order. For the first-order DDIM
update KL is clearly superior (at $M_{\mathrm{eff}}\!=\!128$, KL-DDIM FID\,=\,5.07
vs.\ 5.66 for LC-DDIM, a $10\%$ gap); for the second-order DPM-O2 solver the two
bases fall within a few percent, with Haar marginally ahead (at
$M_{\mathrm{eff}}\!=\!256$, LC-DPM O2 3.94 vs.\ 4.11 for KL-DPM O2). A 3-seed
check places both effects outside seed noise.

\begin{table}[t]
\centering
\caption{KL vs.\ LC samplers on CIFAR-10 PFDiff-790k, FID-50k. KL and LC use
matched effective truncation orders. $S\!=\!20$, $\eta\!=\!0$; see
Supplementary App.~A for LC sampler and evaluation details.}
\label{tab:lc-comparison}
\setlength{\tabcolsep}{4pt}\renewcommand{\arraystretch}{0.95}\small
\begin{tabular}{llr}
\toprule
Sampler & Basis / $M_{\mathrm{eff}}$ & FID-50k \\
\midrule
DDIM        & --- & 6.90  \\
DPM-O2      & --- & 4.61  \\
\midrule
KL-DDIM     & KL / 128 & \best{5.07} \\
LC-DDIM     & LC / 128 ($N\!=\!6$) & 5.66 \\
\midrule
KL-DPM O2   & KL / 256 & 4.11 \\
LC-DPM O2   & LC / 256 ($N\!=\!7$) & \best{3.94} \\
\bottomrule
\end{tabular}
\end{table}

\paragraph{KL vs.\ LC: basis meets solver order}
The KL expansion is the \emph{unique} $L^2$-optimal basis~\cite{loeve1978}: its
eigenfunctions $\psi_m(t)$ are adapted to the SDE covariance through $c_m(t)$, so truncation
drops the least energetic modes first, whereas the Haar basis spreads approximation power
uniformly across dyadic scales regardless of the schedule. This optimality is visible at
first order, where KL-DDIM beats LC-DDIM. At second order (DPM-O2) the solver's own
discretization error dominates the residual basis gap and the two bases become comparable
(Haar marginally ahead). We therefore adopt KL for its $L^2$-optimality and its consistent
first-order advantage, rather than for a universal dominance over Haar.

Additional $S$-sweep curves and representative sample grids are reported
in Supplementary Apps.~B and~C.

\section{Conclusion}
\label{sec:conclusion}

A pretrained diffusion model operates below its prescribed noise level---a counterintuitive
mismatch with the matched-marginal assumption of diffusion. We addressed it with
KL-Diffusion: truncating the Karhunen--Lo\`eve expansion of the Brownian driver yields a
finite-dimensional forward process, a KL training loss, and a family of KL samplers, all
governed by a single truncation order~$M$ that dials the noise level independently of the
step count~$S$.

The central object is the \emph{effective mode count}~$M^*$. A finite-capacity denoiser, fit
on finitely many samples, behaves as the MMSE denoiser at a reduced level
$\seff(t)<\sigma(t)$; a closed-form $2$-Wasserstein analysis then puts a unique interior
optimum on any pointwise-monotone family of reduced schedules, and $M^*$ is read off a
one-dimensional sweep. This single quantity unifies the two uses of the framework. On the
\emph{sampling} axis, KL samplers probe $M^*$ on fixed checkpoints and improve them without
retraining---across CIFAR-10, CelebA, ImageNet~$64\!\times\!64$ and Stable Diffusion v1.4,
and across formula-based (DDIM, DPM-Solver) and learned-predictor (AMED) solvers---with the
optimal order varying but the U-shape invariant. On the \emph{training} axis, a controlled
CIFAR-10 study shows that training-side and sampling-side sweeps recover the same~$M^*$, and
that training at $M'\!=\!M^*$ gives the best-matched pipeline ($2.8\times$ faster convergence
and lower final FID); iterating the match defines a descending-$M^*$ procedure, reported as
an empirical heuristic.

\paragraph*{Limitations and open questions.}
Our generative evaluation centres on FID (complemented by PSNR/SSIM/LPIPS for the inverse
problem); most reported values are single-run, with \Cref{tab:train-scratch} giving the
three-seed spread, so our conclusions rest on the shape of full sweeps rather than on
isolated pairwise differences. The Gaussian analysis controls second moments only, and $M^*$
is located by a per-configuration sweep. Open questions include a convergence analysis of the
descending-$M^*$ procedure, why the swept order depends on the sampler and why the basis
advantage narrows at higher solver order, multi-time KL training objectives, and extensions
to other (and non-SDE) generative processes.

\bibliographystyle{IEEEtran}

\begin{thebibliography}{35}

\bibitem{hoDenoisingDiffusionProbabilistic2020}
J.~Ho, A.~Jain, and P.~Abbeel,
``Denoising diffusion probabilistic models,''
in \emph{Proc. NeurIPS}, 2020.

\bibitem{songScoreBasedGenerativeModeling2021}
Y.~Song, J.~Sohl-Dickstein, D.~P.~Kingma, A.~Kumar, S.~Ermon, and B.~Poole,
``Score-based generative modeling through stochastic differential equations,''
in \emph{Proc. ICLR}, 2021.

\bibitem{songDenoisingDiffusionImplicit2021}
J.~Song, C.~Meng, and S.~Ermon,
``Denoising diffusion implicit models,''
in \emph{Proc. ICLR}, 2021.

\bibitem{lu2022dpm}
C.~Lu, Y.~Zhou, F.~Bao, J.~Chen, C.~Li, and J.~Zhu,
``DPM-Solver: A fast ODE solver for diffusion probabilistic model sampling
in around 10 steps,''
in \emph{Proc. NeurIPS}, 2022.

\bibitem{baoAnalyticDPMAnalyticEstimate2022}
F.~Bao, C.~Li, J.~Zhu, and B.~Zhang,
``Analytic-DPM: An analytic estimate of the optimal reverse variance in
diffusion probabilistic models,''
in \emph{Proc. ICLR}, 2022.

\bibitem{karrasElucidatingDesignSpace2022}
T.~Karras, M.~Aittala, T.~Aila, and S.~Laine,
``Elucidating the design space of diffusion-based generative models,''
in \emph{Proc. NeurIPS}, 2022.

\bibitem{song2023consistency}
Y.~Song, P.~Dhariwal, M.~Chen, and I.~Sutskever,
``Consistency models,''
in \emph{Proc. ICML}, 2023.

\bibitem{liuFlowStraightFast2022b}
X.~Liu, C.~Gong, and Q.~Liu,
``Flow straight and fast: Learning to generate and transfer data with
rectified flows,''
in \emph{Proc. ICLR}, 2023.

\bibitem{salimans2022progressive}
T.~Salimans and J.~Ho,
``Progressive distillation for fast sampling of diffusion models,''
in \emph{Proc. ICLR}, 2022.

\bibitem{hoogeboom2023blurring}
E.~Hoogeboom and T.~Salimans,
``Blurring diffusion models,''
in \emph{Proc. ICLR}, 2023.

\bibitem{bansal2023cold}
A.~Bansal \emph{et al.},
``Cold diffusion: Inverting arbitrary image transforms without noise,''
in \emph{Proc. NeurIPS}, 2023.

\bibitem{yoon2023levy}
E.~B.~Yoon, K.~Park, S.~Kim, and S.~Lim,
``Score-based generative models with L\'evy processes,''
in \emph{Proc. NeurIPS}, 2023.

\bibitem{nobis2024fractional}
G.~Nobis \emph{et al.},
``Generative fractional diffusion models,''
in \emph{Proc. NeurIPS}, 2024.

\bibitem{hu2022lora}
E.~J.~Hu \emph{et al.},
``LoRA: Low-rank adaptation of large language models,''
in \emph{Proc. ICLR}, 2022.

\bibitem{loeve1978}
M.~Lo\`eve,
\emph{Probability Theory II}, 4th~ed. Springer, 1978.

\bibitem{pavliotisStochasticProcessesApplications2014}
G.~A.~Pavliotis,
\emph{Stochastic Processes and Applications}. Springer, 2014.

\bibitem{oksendal2003}
B.~\O{}ksendal,
\emph{Stochastic Differential Equations}, 6th~ed. Springer, 2003.

\bibitem{anderson1982}
B.~D.~O.~Anderson,
``Reverse-time diffusion equation models,''
\emph{Stochastic Process. Appl.}, vol.~12, no.~3, pp.~313--326, 1982.

\bibitem{ghanem2003stochastic}
R.~G.~Ghanem and P.~D.~Spanos,
\emph{Stochastic Finite Elements: A Spectral Approach}, rev.~ed.
Dover, 2003.

\bibitem{schwab2006karhunen}
C.~Schwab and R.~A.~Todor,
``Karhunen--Lo\`eve approximation of random fields by generalized fast
multipole methods,''
\emph{J. Comput. Phys.}, vol.~217, pp.~100--122, 2006.

\bibitem{heusel2017gans}
M.~Heusel, H.~Ramsauer, T.~Unterthiner, B.~Nessler, and S.~Hochreiter,
``GANs trained by a two time-scale update rule converge to a local
Nash equilibrium,''
in \emph{Proc. NeurIPS}, 2017.

\bibitem{wang2025pfdiff}
G.~Wang, Y.~Cai, L.~Li, W.~Peng, and S.-Z.~Su,
``PFDiff: Training-free acceleration of diffusion models combining past
and future scores,''
in \emph{Proc. ICLR}, 2025.

\bibitem{liu2015celeba}
Z.~Liu, P.~Luo, X.~Wang, and X.~Tang,
``Deep learning face attributes in the wild,''
in \emph{Proc. ICCV}, 2015.

\bibitem{rombach2022high}
R.~Rombach, A.~Blattmann, D.~Lorenz, P.~Esser, and B.~Ommer,
``High-resolution image synthesis with latent diffusion models,''
in \emph{Proc. CVPR}, 2022.

\bibitem{ho2022classifier}
J.~Ho and T.~Salimans,
``Classifier-free diffusion guidance,''
in \emph{NeurIPS Workshop on Deep Generative Models}, 2022.

\bibitem{hangEfficientDiffusionTraining2023}
T.~Hang \emph{et al.},
``Efficient diffusion training via Min-SNR weighting strategy,''
in \emph{Proc. ICCV}, 2023.

\bibitem{linCommonDiffusionNoise2024}
S.~Lin, B.~Liu, J.~Li, and X.~Yang,
``Common diffusion noise schedules and sample steps are flawed,''
in \emph{Proc. WACV}, 2024.

\bibitem{kawarSNIPSSolvingNoisy2021}
B.~Kawar, G.~Vaksman, and M.~Elad,
``SNIPS: Solving noisy inverse problems stochastically,''
in \emph{Proc. NeurIPS}, 2021.

\bibitem{zhangDAPS2025}
B.~Zhang, W.~Chu, J.~Berner, C.~Meng, A.~Anandkumar, and Y.~Song,
``Improving diffusion inverse problem solving with decoupled noise annealing,''
in \emph{Proc. CVPR}, 2025.

\bibitem{lu2022dpmpp}
C.~Lu, Y.~Zhou, F.~Bao, J.~Chen, C.~Li, and J.~Zhu,
``DPM-Solver++: Fast solver for guided sampling of diffusion
probabilistic models,''
\emph{Mach. Intell. Res.}, vol.~22, no.~4, pp.~730--751, 2025.

\bibitem{zhou2024amed}
D.~Zhou, D.~Wang, C.~Li, J.~Zhu, and J.~Ye,
``Fast ODE-based sampling for diffusion models in around 5 steps,''
in \emph{Proc. CVPR}, 2024.

\bibitem{deng2009imagenet}
J.~Deng, W.~Dong, R.~Socher, L.-J.~Li, K.~Li, and L.~Fei-Fei,
``ImageNet: A large-scale hierarchical image database,''
in \emph{Proc. CVPR}, 2009.

\bibitem{caoExploringOptimalChoice2023}
Y.~Cao, J.~Chen, Y.~Lu, and E.~G.~Tabak,
``Exploring the optimal choice for generative processes in diffusion models:
Ordinary vs stochastic differential equations,''
\emph{Trans. Mach. Learn. Res.}, 2024.

\bibitem{ningExposureBiasElucidating2024}
M.~Ning, M.~Li, J.~Su, A.~A.~Salah, and I.~O.~Ertugrul,
``Elucidating the exposure bias in diffusion models,''
in \emph{Proc. ICLR}, 2024.

\bibitem{sabourAlignYourSteps2024}
A.~Sabour, S.~Fidler, and K.~Kreis, ``Align Your Steps: Optimizing Sampling Schedules in
Diffusion Models,'' in \emph{Proc. ICML}, 2024.

\bibitem{elementaryScheduling2026}
Q.~Sun, H.~V.~Poor, and W.~Zhang,
``A Gaussian perspective for distributional discrepancy in generative diffusion
models,''
\emph{arXiv:2601.13602}, 2026.

\bibitem{wangVastola2024}
B.~Wang and J.~J.~Vastola,
``The unreasonable effectiveness of Gaussian score approximation for diffusion
models and its applications,''
\emph{Transactions on Machine Learning Research}, 2024.

\bibitem{hurault2024}
S.~Hurault, T.~Moreau, M.~Terris, and G.~Peyr\'e,
``From score matching to diffusion: a fine-grained error analysis in the
Gaussian setting,''
\emph{arXiv:2503.11615}, 2025.

\bibitem{vincent2011}
P.~Vincent,
``A connection between score matching and denoising autoencoders,''
\emph{Neural Computation}, vol.~23, no.~7, pp.~1661--1674, 2011.

\bibitem{efron2011}
B.~Efron,
``Tweedie's formula and selection bias,''
\emph{J. Amer. Statist. Assoc.}, vol.~106, no.~496, pp.~1602--1614, 2011.

\end{thebibliography}

\appendices

\section{Verification of Convergence}
\label{app:convergence}

Throughout we adopt the normalization $T=1$ (the general case follows by time
rescaling) and use the exact representation~\eqref{eq:exact} of
\Cref{sec:parseval} together with its derivative eigenfunctions~\eqref{eq:psi},
which at $T=1$ read $\psi_m(t)=\sqrt{2}\cos\!\big((m-\tfrac12)\pi t\big)$, with
coefficients $c_m(t):=\int_0^t h\,\psi_m\,\dd s$. All results extend
coordinate-wise to $\Real^d$ with $\Var[X_t^{(M)}]=\sigM^2(t)\,I_d$.

\begin{theorem}[Convergence of KL-diffusion]\label{thm:convergence-app}
Assume $h \in C^1([0,1])$.  Then:
\begin{enumerate}
\item[(i)] $\Expect[|X_t - X_t^{(M)}|^2]
  = \alpha(t)^2\sum_{m>M} c_m(t)^2 \to 0$ for each $t$.
\item[(ii)] $\sup_{t \in [0,1]} \Expect[|X_t - X_t^{(M)}|^2] \to 0$.
\item[(iii)] $\sup_{t \in [0,1]} \Expect[|X_t - X_t^{(M)}|^2]
  \leq C/M$ where $C = (\max_t \alpha(t))^2 \cdot A^2/\pi^2$
  and $A = \sqrt{2}(\|h\|_\infty + \|h'\|_\infty)$.
\end{enumerate}
\end{theorem}

\begin{proof}
By~\eqref{eq:exact} and~\eqref{eq:XM} the error is
$X_t - X_t^{(M)} = \alpha(t)\sum_{m>M} c_m(t)\,Z_m$, and since the
$\{Z_m\}$ are independent $\mathcal{N}(0,1)$,
$\Expect[|X_t - X_t^{(M)}|^2] = \alpha(t)^2 \sum_{m>M} c_m(t)^2$.

\emph{(i) Pointwise.}
Parseval's theorem gives $\sum_{m\ge1} c_m(t)^2 = \int_0^t h(s)^2\,\dd s < \infty$,
so the tail $\sum_{m>M} c_m(t)^2 \to 0$.

\emph{(ii)--(iii) Uniform rate.}
We bound $c_m$ by integration by parts. With
$\psi_m(s)=\sqrt{2}\cos(\omega_m s)$, $\omega_m=(m-\tfrac12)\pi$
(from~\eqref{eq:psi} at $T=1$), and $u=h(s)$, $\dd v=\cos(\omega_m s)\,\dd s$,
\[
  c_m(t)=\sqrt{2}\Bigl[\tfrac{h(t)\sin(\omega_m t)}{\omega_m}
  -\tfrac{1}{\omega_m}\!\int_0^t\! h'\sin(\omega_m s)\,\dd s\Bigr],
\]
the boundary term at $0$ vanishing. Using $|\sin|\le1$ and $t\le1$, the two
pieces are at most $\|h\|_\infty/\omega_m$ and $\|h'\|_\infty/\omega_m$, so
\begin{equation}\label{eq:cm-bound}
  |c_m(t)|\le\frac{\sqrt{2}\,(\|h\|_\infty+\|h'\|_\infty)}{\omega_m}=\frac{A}{\omega_m},
\end{equation}
with $A:=\sqrt{2}\,(\|h\|_\infty+\|h'\|_\infty)$, for all $t\in[0,1]$ (finite since
$h\in C^1$). Squaring and using
$(m-\tfrac12)^2\ge(m-1)m$, the tail telescopes:
\begin{align*}
  \sum_{m>M}c_m(t)^2
  &\le\frac{A^2}{\pi^2}\sum_{m>M}\frac{1}{(m-\tfrac12)^2}
   \le\frac{A^2}{\pi^2}\sum_{m>M}\Bigl(\tfrac{1}{m-1}-\tfrac{1}{m}\Bigr)\\
  &=\frac{A^2}{\pi^2 M}.
\end{align*}
Multiplying by $\alpha(t)^2\le(\max_t\alpha(t))^2$ and taking the supremum
gives $\sup_t\Expect[|X_t-X_t^{(M)}|^2]\le C/M$ with
$C=(\max_t\alpha(t))^2 A^2/\pi^2$, which is~(iii); (ii) follows since
$C/M\to0$.
\end{proof}

\section{Uniform-Angle Estimate of \texorpdfstring{$\kappa_S$}{kappa\_S}}
\label{app:uniform-angle}
The discretization factor \eqref{eq:kappa} is the supremum of the $S$-fold cosine product at
fixed total sweep $\Phi=\phi(t_S)$,
$\kappa_S=\sup\{\prod_{k=1}^{S}\cos\delta_k:\delta_k\ge0,\ \sum_{k}\delta_k=\Phi\}$. Since
$\log\cos$ is concave on $[0,\tfrac{\pi}{2})$, Jensen's inequality makes the product largest at
equal angles $\delta_k=\Phi/S$, so, using $\log\cos\delta=-\delta^2/2-\delta^4/12-\cdots$,
\begin{equation}\label{eq:kappa-uniform}
\begin{aligned}
  \kappa_S=\cos^S\!\Big(\frac{\Phi}{S}\Big)
  &=\exp\!\Big(-\frac{\Phi^2}{2S}-\frac{\Phi^4}{12S^3}-\cdots\Big)\\
  &=e^{-\Phi^2/(2S)}\big(1+O(S^{-3})\big).
\end{aligned}
\end{equation}
The deficit is $O(1/S)$: doubling the number of steps roughly halves $1-\kappa_S$. For
$\Phi=\pi/2$, $S=20$ this gives $\kappa_S\approx e^{-\pi^2/160}\approx0.94$.

\end{document}